\documentclass{article}
\usepackage{algorithm}
\usepackage[utf8]{inputenc}
\PassOptionsToPackage{numbers, compress}{natbib}
\usepackage[preprint]{neurips_2025}
\usepackage{arydshln}
\usepackage{colortbl,xcolor}
\usepackage{amsmath}
\usepackage{amsmath, amssymb}
\usepackage{tabularx}   
\usepackage{booktabs}   
\usepackage{float} 
\usepackage[utf8]{inputenc} 
\usepackage[T1]{fontenc}    
\usepackage{hyperref}       
\usepackage{url}            
\usepackage{booktabs}       
\usepackage{amsfonts}       
\usepackage{nicefrac}       
\usepackage{microtype}      
\usepackage{xcolor}         
\usepackage{graphicx}
\usepackage[utf8]{inputenc} 
\usepackage[T1]{fontenc}    
\usepackage{hyperref}       
\usepackage{url}            
\usepackage{booktabs}       
\usepackage{amsfonts}       
\usepackage{nicefrac}       
\usepackage{microtype}      
\usepackage{xcolor}         
\usepackage[utf8]{inputenc} 
\usepackage[T1]{fontenc}    
\usepackage{hyperref}       
\usepackage{url}            
\usepackage{booktabs}       
\usepackage{amsfonts}       
\usepackage{nicefrac}       
\usepackage{enumitem}
\usepackage{algorithm}      
\usepackage{algpseudocode}  
\usepackage{amsmath}        
\usepackage{graphicx}
\usepackage{float}
\usepackage{listings}
\floatstyle{ruled}
\newfloat{listing}{tbp}{lop}
\floatname{listing}{Listing}
\usepackage{listings}
\usepackage{subcaption} 
\usepackage{caption}
\usepackage{microtype}      
\usepackage{xcolor}         
\usepackage{xpatch}
\usepackage{tabularx} 
\usepackage{listings}   
\usepackage{xpatch}
\usepackage{booktabs,tabularx,graphicx}
\usepackage[table]{xcolor} 
\title{Law in Silico: Simulating Legal Society with LLM-Based Agents}

\author{
  Yiding Wang$^{1,2\ *}$ \And Yuxuan Chen$^{3}$ \thanks{Equal Contribution.} \AND Fanxu Meng$^1$ \And Xifan Chen$^4$ \And Xiaolei Yang$^5$ \And Muhan Zhang$^{1,6}$ \\
  $^1$Institute for Artificial Intelligence, Peking University \quad
  $^2$Yuanpei College, Peking University \\
  $^3$School of Computing and Data Science,
  The University of Hong Kong \\ $^4$ Individual Researcher \quad $^5$ Law School, Peking University \\ 
  $^6$ State Key Laboratory of General Artificial Intelligence, BIGAI \\
  \texttt{yidingw@stu.pku.edu.cn, ddddennis.chen@gmail.com, muhan@pku.edu.cn}
}

\begin{document}

\maketitle

\begin{abstract}
Since real-world legal experiments are often costly or infeasible, simulating legal societies with Artificial Intelligence (AI) systems provides an effective alternative for verifying and developing legal theory, as well as supporting legal administration. Large Language Models (LLMs), with their world knowledge and role-playing capabilities, are strong candidates to serve as the foundation for legal society simulation. However, the application of LLMs to simulate legal systems remains underexplored. In this work, we introduce \textbf{Law in Silico}, an LLM-based agent framework for simulating legal scenarios with individual decision-making and institutional mechanisms of legislation, adjudication, and enforcement. Our experiments, which compare simulated crime rates with real-world data, demonstrate that LLM-based agents can largely reproduce macro-level crime trends and provide insights that align with real-world observations. At the same time, micro-level simulations reveal that a well-functioning, transparent, and adaptive legal system offers better protection of the rights of vulnerable individuals.
\end{abstract}

\section{Introduction}

\noindent \textit{``We wish to acquire knowledge about a target entity T. But T is not easy to study directly. So we proceed indirectly. Instead of T we study another entity M, the `model' ...''}  

\hfill --- \citet{gilbert2018simulating}

The analysis of law has traditionally been carried out through analytic methods, which rely heavily on theoretical frameworks and retrospective analyses~\citep{aikenhead1999exploring}. While this method has proven valuable, it restricts our ability to understand the dynamic and evolving nature of legal systems. Legal systems, like any social system, are profoundly shaped by the interactions among individuals, institutions, and the broader societal context. As a result, distributed AI systems present a promising alternative to traditional legal analysis: by simulating legal systems, we can observe how laws affect individual behavior and societal dynamics in ways that would be difficult to study directly in the real world.

Recent advancements in large language models (LLMs)~\cite{openai2024gpt4technicalreport}, which are trained on vast and diverse real-world corpora, have demonstrated impressive capabilities in language modeling and generalization across a wide range of downstream tasks. These models are capable of understanding the languages, cultures, and societies of different countries~\cite{Shah2024HowWDA}, essentially providing a probabilistic model of the world. LLM-based agents~\cite{Wang_2024} leverage the power of context, transforming the general world distributions learned by these models into a specific, individual-focused profile distribution~\cite{Bui2025MixtureofPersonasLMA}. This enables the models to exhibit strong role-playing capabilities, where decisions are based on the agent's background and situational context. By integrating these capabilities, LLM-based agent systems~\cite{Luo2025LargeLMA} provide a powerful and scalable approach to simulating legal societies, offering a unique way to model the complexities of legal decision-making and agent interactions.

\begin{figure*}[t]
    \centering
    \vspace{-20pt}
    \includegraphics[width=\textwidth]{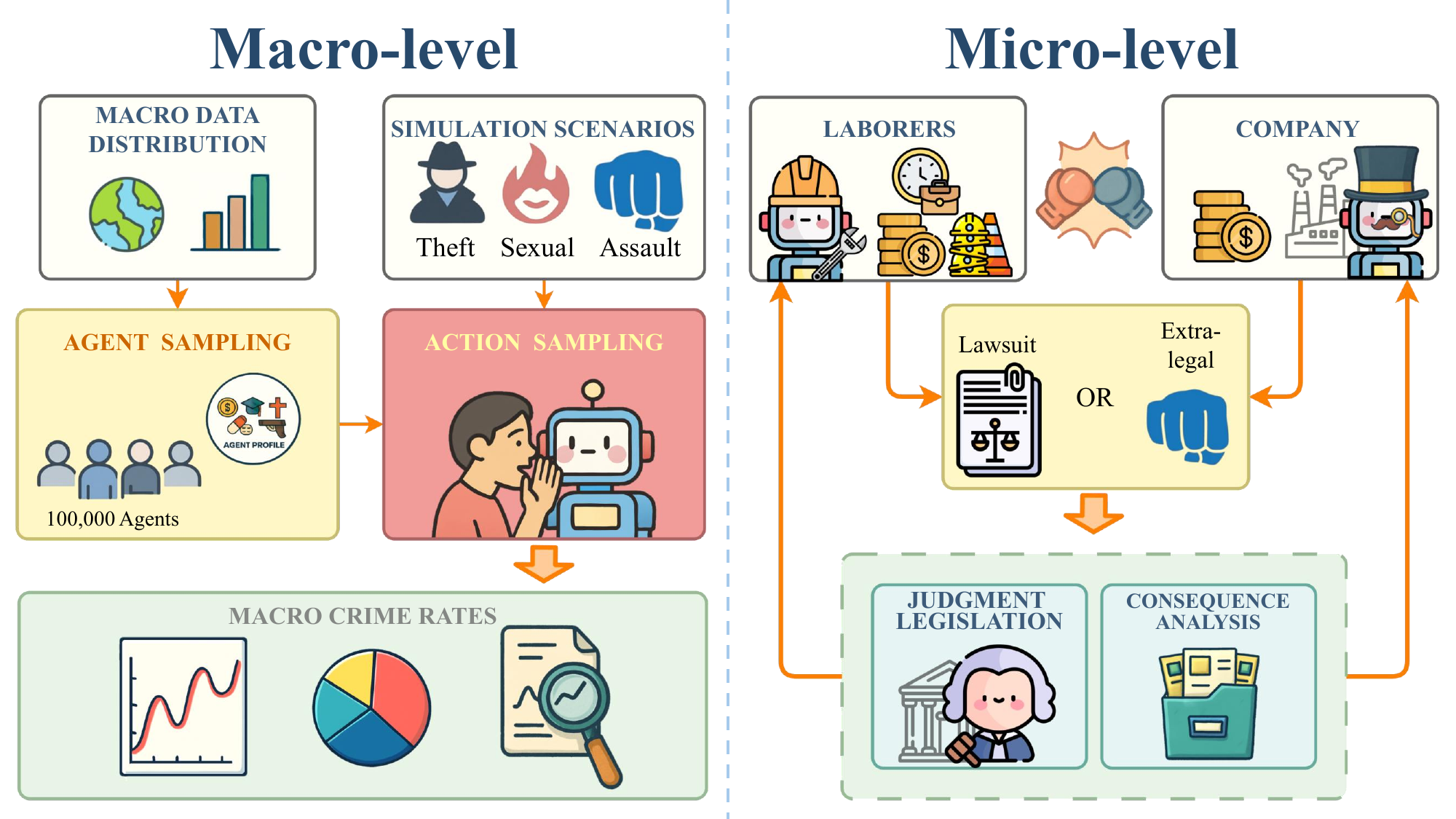}
    \caption{Overview of the \textbf{Law in Silico} framework. The left sub-figure represents the macro-level simulation, where broad societal data and legal regulations collectively
    inform agent behaviors. The right sub-figure illustrates the micro-level simulation, which focuses on multi-turn interactions between individual agents, moderated by an LLM-powered Game Master, to simulate legal judgements and institutional dynamics.}
    \label{fig:framework}
\end{figure*}

While prior studies explore general social simulations~\citep{piao2025agentsocietylargescalesimulationllmdriven} or specific legal tasks like judicial decision-making~\citep{he2024agentscourtbuildingjudicialdecisionmaking,sun2024lawluomultiagentcollaborativeframework}, they do not examine how macro-level societal factors influence individual choices, nor how agent interactions contribute to the emergent evolution of legal systems.
Our work bridges this gap by introducing \textbf{Law in Silico}, a simulation framework that comprises: (1) \textbf{\textit{Hierarchical Legal Agent Modeling}:} We employ hierarchical sampling from real-world statistical distributions to approximate the multivariate characteristics relevant to legal decision-making and societal behavior. (2) \textbf{\textit{Scenario-Based Decision-Making}:} Our framework supports both large-scale, single-shot decision-making simulations and interactive, multi-turn simulations in specific scenarios (\textit{e.g.,} crime exposure or conflicts of interest). (3) \textbf{\textit{Legal System}:} We enable the flexible initialization and modification of laws, implementing evolving legislative mechanisms while considering factors such as corruption in the judiciary to build a more realistic legal system. Through macro-level simulations of illegal scenarios (theft, assault, and sex trade), we find that the crime rates simulated by our system largely align with real-world macro crime rates, while also revealing consistent patterns across various crime-related factors. In our micro-level experiments, we compared different legal system configurations in scenarios involving labor exploitation and anti-exploitation. We surprisingly reproduced the ``cat-and-mouse'' dynamics observed in the real world. Moreover, results from controlled experiments demonstrate the importance of a well-functioning, transparent, and efficient legal system in protecting the welfare of vulnerable populations.

Our main contributions can be summarized as follows: 
\begin{itemize}
    \item \textbf{A Comprehensive Framework for Legal Simulation}: We propose \textbf{Law in Silico}, a novel simulation framework that combines hierarchical legal agent modeling based on real-world statistical data, diverse scenario-based decision-making (both single-shot and interactive), and dynamic legal system mechanisms, including realistic factors such as judicial corruption.
    \item \textbf{Empirical Validation of LLM-based Legal Simulations}: We demonstrate that our LLM-based agent system can realistically replicate macro-level crime trends, with simulated crime rates closely aligning with real-world data and capturing key influencing variables.
    \item \textbf{Insights into Legal System Design and Evolution}: Our micro-level experiments provide crucial insights into the characteristics of effective legal systems and how agent interactions can drive the evolution of legal frameworks, particularly in protecting vulnerable groups.
\end{itemize}

\section{Related Work}

\paragraph{LLM-Based Agents}

The emergence of large language models (LLMs) has revolutionized various fields, demonstrating remarkable capabilities in natural language understanding, generation, and complex reasoning~\cite{Wei2022ChainOTA}. Beyond traditional tasks, LLMs are increasingly leveraged to create autonomous agents~\cite{Wang2023ASOA} capable of simulating human-like behavior and interactions within diverse environments. Early work in this area primarily focused on developing LLM-driven agents for tasks such as planning, problem-solving, and conversational AI, with notable examples including ReAct~\cite{yao2023reactsynergizingreasoningacting}, BabyAGI, and Voyager~\cite{wang2023voyageropenendedembodiedagent}, all of which integrate reasoning, action, and exploration for autonomous decision-making. Subsequent research has explored architectural improvements and prompting strategies to enhance agent autonomy and decision-making in open-ended environments. For example, LLMob~\cite{jiawei2024large} developed an LLM agent framework for personal mobility generation, aligning LLMs with real-world activity patterns to simulate human-like mobility behaviors. Moreover, the Desire-driven Autonomous Agent (D2A)~\cite{wang2024simulating} autonomously selects tasks driven by multi-dimensional desires, such as social interaction and self-fulfillment, enabling more adaptive and human-like behavior. Despite the numerous explorations of agent-based models, there is still a gap in research regarding the modeling of agents in the context of large-scale legal societies. Whether LLM-based agents can accurately express individual profiles in criminal contexts—and make decisions consistent with real-world macro and micro-level trends—remains an open question.

\paragraph{Legal Society Simulation}

Simulating legal societies can provide valuable insights into how laws influence behavior and shape societal dynamics. These simulations allow researchers to test legal frameworks, understand law enforcement mechanisms, and predict legal outcomes in scenarios that are difficult or costly to explore in the real world~\cite{aikenhead1999exploring}. Early works in legal simulations often relied on rule-based knowledge systems~\cite{sergot1986british} to model and simulate legal processes. Although LLM-based social simulations have been explored in works such as Smallville~\cite{park2023generativeagentsinteractivesimulacra} and AgentSociety~\cite{piao2025agentsocietylargescalesimulationllmdriven}, large-scale legal society simulations are still rare. Most existing work focuses on societal behaviors without incorporating complex legal systems at scale. Recent legal agent-based simulations using LLMs include AgentsCourt~\cite{he2024agentscourtbuildingjudicialdecisionmaking}, which simulates courtroom processes with judge and lawyer agents, and Agents on the Bench~\cite{jiang2024agentsbenchlargelanguage}, which models judicial decision-making with judge and juror agents to improve fairness and accuracy. LawLuo~\cite{sun2024lawluomultiagentcollaborativeframework} simulates legal consultations with multi-agent interactions mimicking law firm operations, while MASER~\cite{yue2025multiagentsimulatordriveslanguage} generates data for legal training via agent-based simulations. However, these works mostly focus on specific legal domains, with little research on large-scale legal societies that integrate both macro and micro-level legal dynamics. Our work bridges this gap by modeling agents in legal societies that simulate the macro-level impact of law on individuals, while also supporting the complex, dynamic processes of legislation and law enforcement at the micro-level.

\section{Law in Silico}

In this Section, we introduce \textbf{Law in Silico}, an LLM-based legal simulation framework. Our framework places large language models (LLMs) at the core of agent, legislative and judicial decision-making, aiming to simulate key dimensions of legal societies—including individual-level crime propensity, rights-protection dynamics, and the operation of enforcement and legislative mechanisms. We use real-world societal statistics to construct agent profiles, allowing the simulation to reflect social heterogeneity relevant to crime and legal behavior. As illustrated in Figure~\ref{fig:framework}, the framework supports both macro-level statistical simulations—examining how aggregate indicators such as income, education, and drugs influence crime rates—and micro-level simulations that model how legal institutions affect decision-making and welfare in bilateral conflict scenarios. The framework consists of three main components: \textit{Hierarchical Legal Agent Modeling}, \textit{Scenario-Based Decision-Making}, and the \textit{Legal System}.

\subsection{Hierarchical Legal Agent Modeling}

Research in criminology and legal studies has identified a wide range of factors that influence an individual's propensity to commit a crime~\citep{landau1997crime}. To build realistic legal agents, we incorporate these factors into the internal profiles of LLM-based agents, such that each agent carries a representation of their socioeconomic background, social conditioning, and legal perceptions. These internalized traits influence how agents perceive risk, make decisions, and finally interpret legal constraints in downstream scenarios.

We model three primary categories of factors: (1) \textbf{\textit{socioeconomic factors}}, such as poverty and inequality, unemployment, and disparities in educational attainment, which shape the agent's perceived opportunity structure and life incentives; (2) \textbf{\textit{social environment}}, including religious affiliation, societal background (\textit{e.g.,} community cohesion), and exposure to drugs or gang influence, which inform the agent's moral reasoning and behavioral norms; and (3) \textbf{\textit{legal factors}}, such as perceived punishment severity and law enforcement effectiveness, which are integrated into the agent's mind as a \textbf{\textit{punishment impression}}—a subjective mental model of legal risk and deterrence.

In macro-level simulations, we ground these attributes using official statistical data, enabling us to reflect population-level variations across different societal contexts. Rather than sampling each factor independently, we adopt a \textit{hierarchical sampling} strategy that accounts for correlations among variables—for example, the relationship between income and education, gender and employment, or age and likelihood of drug or gang involvement. This approach allows us to approximate realistic population distributions and preserve structural dependencies observed in empirical data. In micro-level simulations, agent profiles can be further enriched with individualized narratives, such as occupational roles or personality traits, to support fine-grained decision modeling. These personalized elements help simulate the diversity of motivations and constraints that real individuals bring into legal interactions.

\subsection{Scenario-Based Decision Making}

Our framework supports two complementary modes of decision simulation: large-scale, single-shot decision making for macro-level analysis, and multi-turn, interactive decision making among multiple agents for micro-level simulations. In both settings, agent behavior is guided by a scenario context that integrates the agent's profile with
the surrounding environment or situation. These scenarios serve as the basis for generating realistic behavioral responses under legal or moral pressure.

In macro-level simulations, we design situational prompts that expose agents to potential crime-inducing environments, accompanied by a predefined set of action options, typically including both legal and illegal choices. Since it is infeasible to simulate an agent's continuous experience in the real world, we approximate crime tendencies by observing agent decisions in these representative high-risk contexts. Agent profiles are combined with scenario descriptions to form descriptive decision contexts, which are then processed in batch by an optimized LLM engine (\textit{e.g.,} vLLM) to efficiently generate decisions across thousands of agents. The resulting choice distributions are aggregated for statistical analysis, such as estimating scenario-specific crime rates.

In micro-level simulations, agents engage in sequential, multi-agent interactions under a shared and partially observed scenario. Each agent selects actions turn by turn, based on its own profile, the evolving environment, and the behavior of other agents. To handle the open-ended nature of LLM-generated actions, we introduce a \textit{Game Master} (GM) module—conceptually inspired by tabletop role-playing games—that interprets actions, applies relevant legal or institutional rules, and determines the resulting consequences. The GM is also LLM-powered and plays a key role in managing state transitions and ensuring consistent feedback throughout the simulation.

\subsection{Legal System}

Our framework incorporates a legal system that governs agent behavior and rights enforcement. It consists of four components: a body of law, a legislative mechanism, a judicial mechanism, and an enforcement mechanism. We can include real-world statutes—\textit{e.g.,} criminal codes from existing jurisdictions—or synthetic rules specifically designed to control experimental conditions and isolate causal effects. While legal rules influence decision-making in both macro- and micro-level simulations, the legislative, judicial and enforcement mechanisms primarily operate in the micro setting, where they shape agents’ rights-protection behavior and institutional trust. 

The legislative module is LLM-driven and functions as a rule-evolving system. At predefined intervals, it collects cases or lawsuits initiated by agents—especially those triggered by a judicial ruling where no existing law applies—and evaluates whether the existing legal framework sufficiently protects their interests. This evaluation may lead to the creation, modification, or removal of legal rules. The system supports two initialization modes: starting from an empty legal corpus to simulate legal emergence, or from an existing legal base to study its evolution.
The judicial and enforcement modules are also powered by LLMs. The judicial module determines whether agent actions violate applicable laws, operating on the principle of \textit{Nulla poena sine lege} (no penalty without a law), while the enforcement module determines what penalties should be imposed. 
In addition, we model institutional fairness by introducing a corruption factor, which probabilistically alters the result of the judgment. This allows the system to simulate not only normative legal consequences but also deviations arising from biased or opaque enforcement environments.

\section{Experiments}

We conduct experiments to evaluate whether our \textbf{Law in Silico}  simulation framework can capture real-world behavioral trends and reflect institutional effects of the legal system. We divide our evaluation into two parts: macro-level simulations focus on statistical patterns of agent behavior under varying social and legal configurations, while micro-level simulations explore how legal mechanisms influence agent-level outcomes in interactive conflict scenarios.

\subsection{Macro-Level Simulations}

\paragraph{Experimental Setup.}  

To examine aggregate behavioral patterns across different societal configurations, we collect macro-level economic and demographic data from four representative countries—two developed and two developing—selected to ensure regional diversity and data availability (see Appendix~\ref{app:countries}). Agent profiles are generated using hierarchical sampling conditioned on country-specific distributions of key attributes, including age, gender, education, religion, income, drug use, and gang involvement. In addition to the attributes, each profile includes a \textit{society background} describing the country’s level of development, social cohesion, firearm regulation policies, and immigration patterns, providing detailed background for decision modeling.

We design three decision scenarios for simulation. Two—\textit{theft} (shoplifting from a luxury retail store) and \textit{assault} (barroom bottle attack)—are universally recognized as crimes in all four countries. The third scenario, \textit{sex trade}, reflects jurisdictional differences: it is illegal in Country A (most regions) and Country C (nationwide), while considered legal under regulated conditions in Country B and Country D. Each scenario presents agents with two legal options and one illegal option. To systematically study behavioral responses, we conduct the following experiments:
(1) \textbf{Effect of Legal Deterrence}: We vary the severity of perceived punishment impression across six levels (0–5), ranging from no perceived consequence to extremely severe punishment. We also include a baseline condition without any punishment impression to examine the model’s prior behavior without explicit deterrence cues.  
(2) \textbf{Sociodemographic Influence}: We run controlled comparisons to investigate how attributes such as religion, nationality, and immigrant status affect the likelihood of committing crimes across different legal and cultural settings.

We sample $10,000$ agents and evaluate each in a single-shot setting using temperature-1 decoding. Each agent's behavior is generated based on their profile and the given scenario. Our main experiments use two strong open-source instruction-tuned models—LLaMA3.3-70B-Instruct~\cite{grattafiori2024llama3herdmodels} and Qwen2.5-72B-Instruct~\cite{qwen2025qwen25technicalreport}. The resulting decisions are aggregated to estimate scenario-specific crime rates, which are then compared to real-world statistics if available.

\begin{table*}[t]
\centering
\caption{Real-world and simulated crime rates across four countries (A–D) and three crime types. Simulated values are aggregated from single-shot LLM decisions across 10,000 agents. The real-world crime rates are the most recent available data. Note that for prostitution, no specific crime rate data is available.}
\begin{tabular}{lcccc}
\toprule
& \multicolumn{2}{c}{\textbf{Developed Countries}} & \multicolumn{2}{c}{\textbf{Developing Countries}} \\
\cmidrule(lr){2-3} \cmidrule(lr){4-5}
\textbf{Crime Type} & \textbf{Country A} & \textbf{Country B} & \textbf{Country C} & \textbf{Country D} \\
\midrule
\multicolumn{5}{l}{\textit{Real-World Crime Rate (per capita)}} \\
Larceny-Theft & 0.0135 & 0.0157 & 0.0001 & 0.0004 \\
Assault       & 0.0026 & 0.0017 & 0.0001 & 0.0003 \\
Prostitution  & Illegal        & Legal          & Illegal        & Legal \\
\midrule
\multicolumn{5}{l}{\textit{LLaMA3.3-70B-Instruct}} \\
Larceny-Theft & 0.0103 (-0.0032) & \textbf{0.0159 (+0.0001)} & 0.0010 (+0.0009) & 0.0241 (+0.0237) \\
Assault       & 0.1199 (+0.1173) & 0.0713 (+0.0696) & 0.3624 (+0.3623) & 0.4100 (+0.4097) \\
Prostitution  & 0.0369  & 0.0538  & 0.0002 & 0.0527 \\
\midrule
\multicolumn{5}{l}{\textit{Qwen2.5-72B-Instruct}} \\
Larceny-Theft & \textbf{0.0116 (-0.0019)} & 0.0179 (+0.0021) & \textbf{0.0003 (+0.0002)} & \textbf{0.0028 (+0.0024)} \\
Assault       & \textbf{0.0025 (-0.0001)} & \textbf{0.0022 (+0.0005)} & \textbf{0.0001 (+0.0000)} & \textbf{0.0217 (+0.0216)} \\
Prostitution  & 0.0148 & 0.0194 & 0.0007  & 0.0461 \\
\bottomrule
\end{tabular}
\label{tab:crime-rates}
\end{table*}

\begin{figure*}[t]
    \centering
    \includegraphics[width=\textwidth]{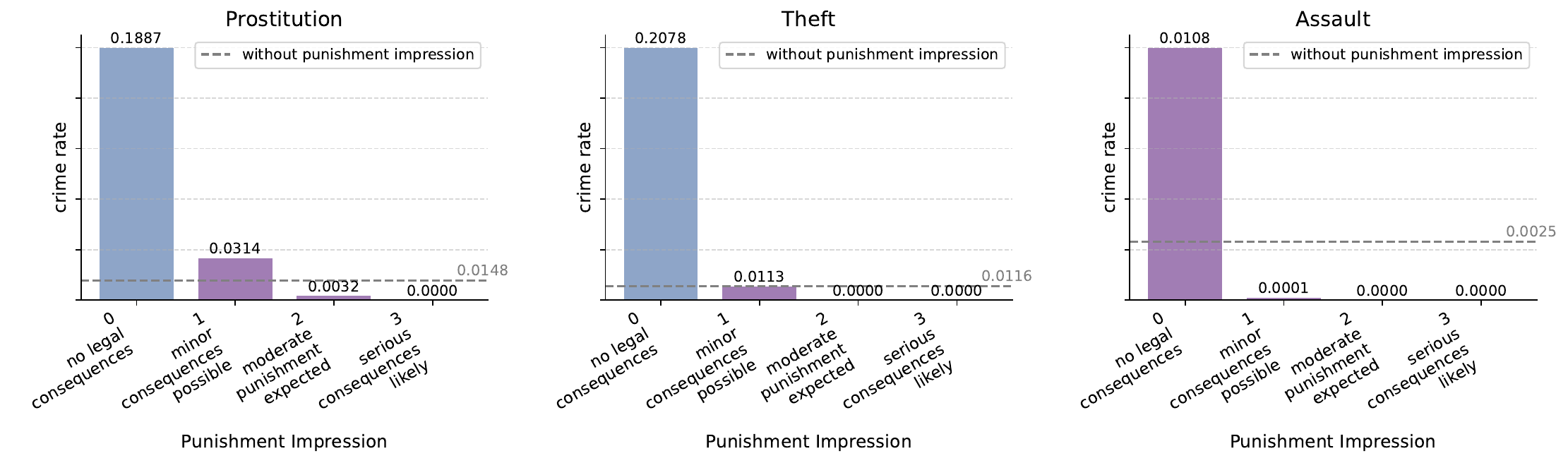}
    \caption{Experiments conducted with the Qwen2.5-72B-Instruct model showed crime rates across punishment impression levels for prostitution, theft, and assault (Country A). Gray dashed lines indicate the baseline rate without punishment impression. Purple bars represent the simulated crime rates that are closest to the baseline (agents are not provided with a punishment impression).}
    \label{fig:punishment-rates}
\end{figure*}

\paragraph{Results and Analysis.}

Table~\ref{tab:crime-rates} presents a comparison between real-world crime rates and the crime rates simulated by LLM-based agents in our framework, where agents are given only their profile and scenario descriptions without explicit legal knowledge or punishment information. This setup allows us to observe the model's prior behavioral tendencies based on socio-contextual cues alone. \textbf{Our results show that simulated crime rates largely reflect real-world statistics.} Qwen2.5-72B-Instruct, in particular, demonstrates high fidelity, \textbf{with errors not exceeding 0.002 across Countries A, B, and C.} In contrast, LLaMA3.3-70B tends to overestimate assault rates, though its overall estimates remain within a reasonable range for most countries. One notable trend is that both models consistently predict higher crime rates for developing countries, especially Country D, compared to the officially reported crime rates. We believe this discrepancy is not due to model failure, but rather reflects differences in how crime is recorded. In developing countries, official crime rates are often based on reported cases, which may underrepresent actual crime, especially for offenses like larceny or assault. Factors such as rural populations, limited police presence, and lower case registration rates contribute to underreporting, making the simulated crime rates appear higher. These findings suggest that, even without explicit legal guidance, LLM-based agents can capture meaningful behavioral patterns aligned with macro-level social and legal structures. At the same time, they reveal the gaps between reported crime data and the latent crime risk in under-resourced regions.

Figure~\ref{fig:punishment-rates} illustrates how varying levels of punishment impression influence simulated crime rates across the three scenarios in Country A. We observe a clear deterrent pattern: \textbf{when agents are explicitly informed that there are no legal consequences (\textit{e.g.,} “people rarely get in trouble for it”), the simulated crime rates increase sharply across all crime types.} As the severity of punishment impression increases—ranging from minor penalties to multi-year imprisonment—the crime rates consistently decline. Notably, when the perceived consequence reaches level 3 (“serious consequences likely”), none of the simulations exhibit a non-zero crime rate for the three scenarios. Compared to the gray dashed baseline line (no punishment impression), we find that \textbf{punishment levels closest to the baseline crime rates tend to align with Country A’s actual legal environment.} For example, in most regions of Country A, a single shoplifting offense may result in minor consequences such as fines (also probably won't be caught), while consensual adult sex work may range from fines to short-term incarceration, depending on jurisdiction. These results suggest that model behavior is sensitive to perceived legal consequences and that alignment between punishment impression and actual deterrent policies yields crime rates that closely reflect real-world statistics. Similar deterrent effects are observed in other countries as well, though with slight variations due to differences in legal structure and base crime rates (see Appendix~\ref{app:punishlevel_results}).

Additionally, we observe a strong alignment between simulated macro-level crime rates and real-world correlations with various factors.  \textbf{Factors such as young, low educational attainment, low income, male, drug use, and gang involvement are consistently associated with higher simulated crime rates, as seen in real-world data.}  Conversely, \textbf{agents with a clear religious affiliation are perceived by the simulator to exhibit lower crime rates}, reflecting the protective effect often attributed to religious socialization in various societies. Moreover, for countries A and B, which have substantial immigrant populations, we find that \textbf{the immigrant status does not lead the simulator to predict higher crime rates. Instead, it correlates with a higher likelihood of engaging in sex trade (or prostitution).} This suggests that LLMs understand the socio-economic pressures and legal ambiguities that immigrants often face. (More details in Appendix~\ref{app:indicators_results} and ~\ref{app:immigrants}.)

\subsection{Micro-Level Simulations}

\begin{figure*}[t]
    \centering
    \begin{subfigure}{0.32\textwidth}
        \centering
        \includegraphics[width=\linewidth]{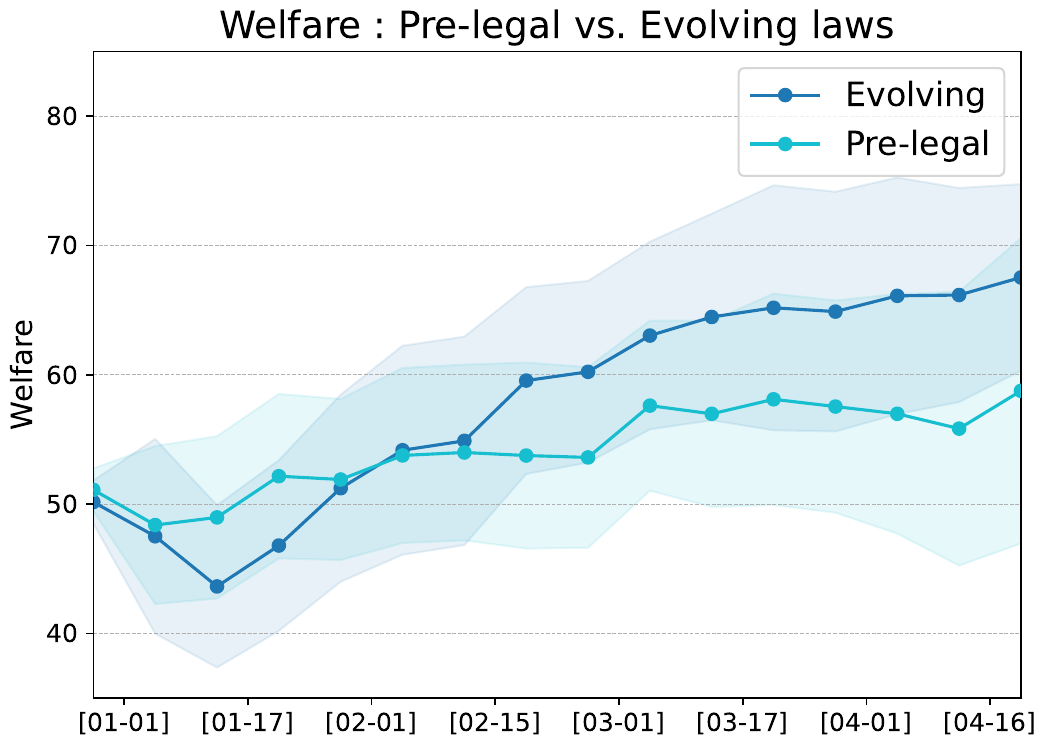}
        \caption{The welfare of laborers under the Pre-Legal and Evolving legal settings}
        \label{fig:Pre-legal}
    \end{subfigure}
    \hfill
    \begin{subfigure}{0.32\textwidth}
        \centering
        \includegraphics[width=\linewidth]{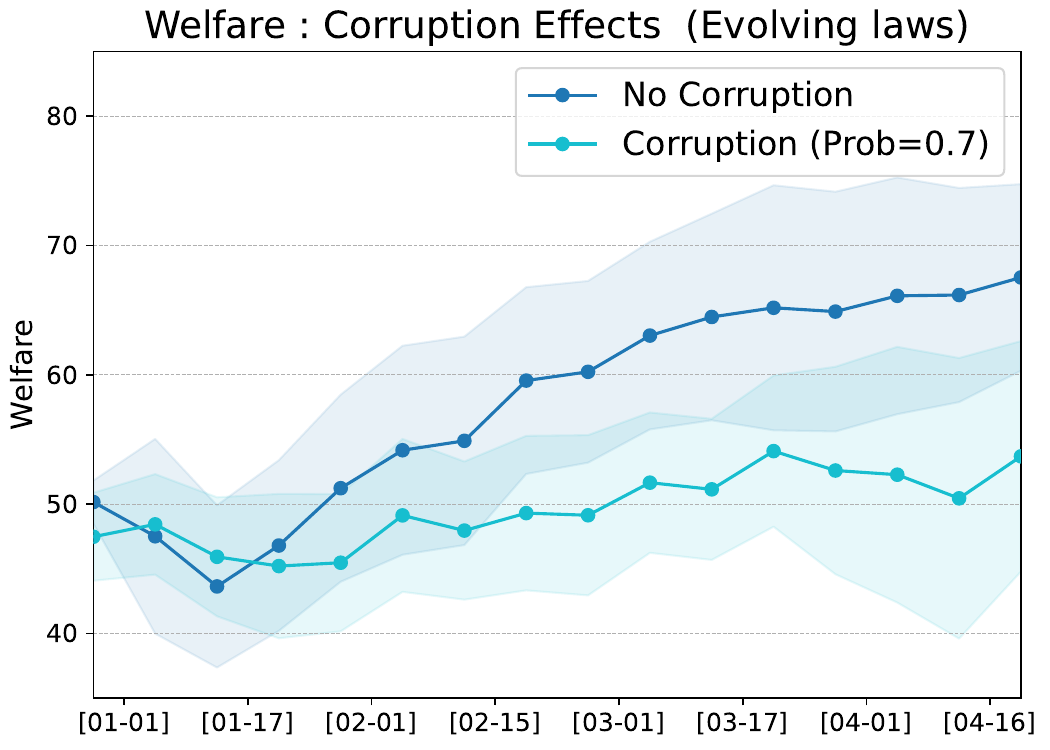}
        \caption{The effects of corruption on laborers’ welfare under the evolving legal system.}
        \label{fig:Corruption}
    \end{subfigure}
    \hfill
    \begin{subfigure}{0.32\textwidth}
        \centering
        \includegraphics[width=\linewidth]{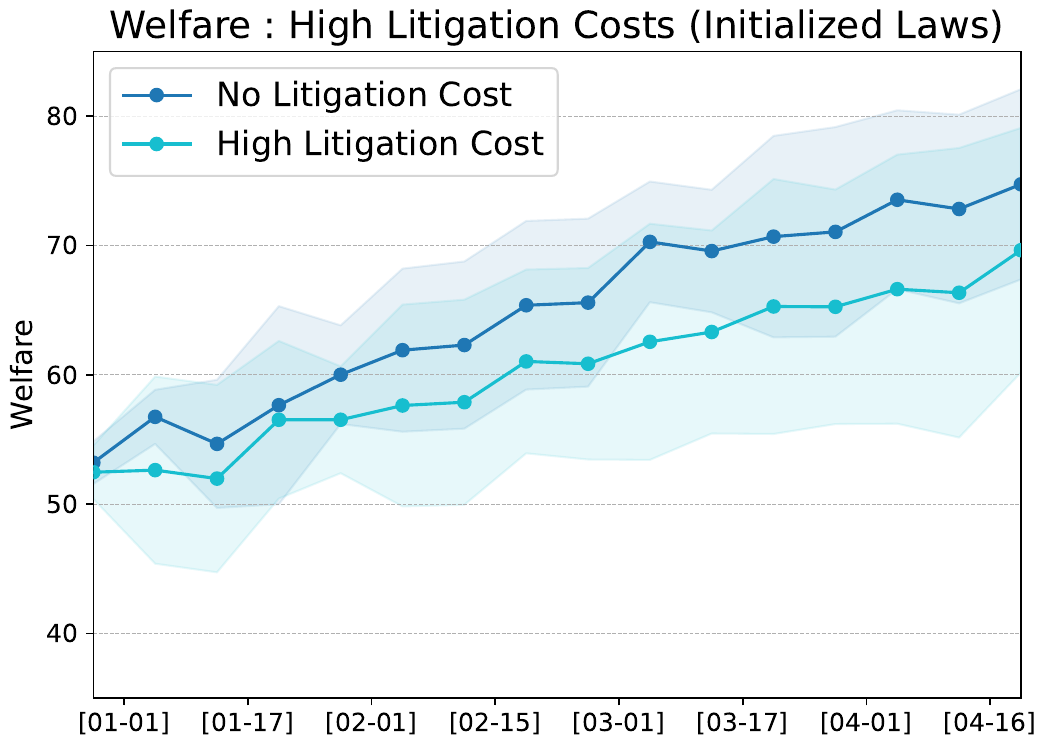}
        \caption{The effect of litigation costs on laborers’ welfare under the initialized law setting.}
        \label{fig:Litigation}
    \end{subfigure}
    \caption{Welfare over time in the Micro-Level Simulation experiments. The solid lines represent the mean welfare, and the shaded areas represent \(\pm1\) standard deviation around the mean.}
    \label{fig:micro_main}
\end{figure*}

\paragraph{Experimental Setup.} To observe the impact of legal systems in conflict scenarios, we design a game-theoretic scenario involving a company and its three laborers with different personalities, situated within a virtual, isolated town. In this scenario, the company's primary objective is to maximize its \textbf{profit} and \textbf{capital}, motivating actions such as reducing hourly wages or proposing mandatory overtime. Conversely, the laborers' goal is to maximize their \textbf{welfare}, determined by a weighted sum of cash, total working hours (calculated as a negative indicator), average hourly wage, and the company's safety investment. When their interests are harmed, laborers can take actions to protect themselves, ranging from legal (\textit{e.g.,} filing lawsuits) to extra-legal (\textit{e.g.,} organizing strikes).

To ensure simulation efficiency, the simulation progresses in discrete time steps, with each agent acting twice per month. The entire simulation spans four months, from January 1st to April 30th. Agents' decision-making and legal system operations are handled by the DeepSeek-Chat model for its impressive abilities, with the temperature set to 1.0 to promote diverse strategic behaviors. To ground the simulation in realism, the following initial parameters are set: a living cost of 1500 units per month, an average hourly wage of 30 units, a standard 40-hour workweek, and a monthly safety investment of 500 units. These values are based on real-world data, reflecting income levels and living costs of blue-collar laborers in non-metropolitan areas of developed countries.

We design six comparative experimental settings to investigate several key conditions and run 6 simulations for each. The experiments discussed in the main body include: 1) A comparison between a completely pre-legal environment and an evolving system derived from a legal vacuum. 2) The effects of legal corruption on both stakeholders. 3) The influence of high litigation costs on the laborers. The primary settings analyzed in the main body of this paper are defined as follows. 
\begin{itemize}
    \item \textit{Pre-Legal (Anarchy)}: This represents a state of anarchy where no legal operations occur.
    \item \textit{Evolving Legal System}: This setting serves as our control group in the \textit{Corruption} experiment. It starts with no laws but allows for the emergence and evolution of laws.
    \item \textit{Corruption}: This setting modifies the \textit{Evolving Legal System} by introducing a probability of \(p=0.7\) that any judicial ruling or legislative event favorable to laborers will be overturned and instead favor the company.
    \item \textit{Initialized Legal System}: This setting serves as the control group in the \textit{Litigation Costs} experiment. It starts with a basic legal framework that includes fundamental labor protections and allows for the evolution of laws over time.
    \item \textit{High Litigation Costs}: This setting modifies the Initialized Legal System by requiring laborers to pay litigation fees to file a lawsuit, with the act of filing also counted as absenteeism.
\end{itemize}
A description of the remaining experimental settings can be found in the Appendix~\ref{app:configurations} and ~\ref{app:additional_exp}.

\paragraph{Results and Analysis.} Figure~\ref{fig:micro_main} illustrates how the laborers' welfare changes over time across three comparative simulations. In all simulations, the company agent strictly follows its core objective of capital maximization. Consequently, in the absence of effective legal enforcement, the company identifies exploitative strategies—such as reducing safety investments or suppressing wages—as optimal paths for short-term profit. When legislation addresses specific loopholes (\textit{e.g.,} by increasing penalty fines), the company dynamically recalculates risk and shifts toward alternative, unregulated strategies with positive expected gains. This adaptive behavior generates a persistent \textbf{``cat-and-mouse'' dynamic} between regulation and corporate conduct. The behavior of laborers in the simulations is strongly related to their perceived strength of protection from law, and the transparency and effectiveness of the legal system. \textbf{When they perceive the legal framework as weak, non-transparent, or ineffective, they tend to adopt more aggressive methods to protect their fundamental rights.}

In the experiment comparing the \textit{Pre-Legal} and \textit{Evolving Legal System} settings, \textbf{the average welfare of the three laborers in the Pre-Legal setting is generally lower than in the Evolving Law setting.} A key difference observed is that, since there is no third-party mediation or enforced regulations in \textbf{\textit{Pre-Legal setting}}, laborers primarily engage in negotiation (by those with calm personalities) and protest (by aggressive or opportunistic laborers). When negotiation fails, protest becomes more likely. As shown in Figure~\ref{fig:Pre-legal}, laborers’ welfare is initially slightly higher in the \textit{Pre-Legal} simulation. In the absence of legal protections, workers defend their interests through strong resistance—such as damaging company property, protesting, and striking. To suppress these actions, the company makes minor concessions, resulting in a temporary increase in welfare. However, this recovery is not sustainable. The company continues its attempts to exploit the laborers. To divide the workforce and prevent collective strikes, the company even offers bonuses to specific laborers, such as with a higher hourly overtime wage and safety investment. This undermines laborer unity, as laborers might choose to work individually or negotiate for personal benefits, leading to moderate and unstable welfare levels \((\textbf{58.75} \pm 11.80)\). Over six trials, we observe the company attempting to divide laborers 8 times, and in 7 of these cases, fewer than two laborers continued protesting after the company's intervention. In the \textbf{\textit{Evolving Legal System}} setting, which starts with a legal vacuum, laborers initially try peaceful methods, such as filing lawsuits to warn the company. However, due to the lack of established laws, the company disregards these actions and continues its exploitation. In response, the laborers begin to protest and strike. When the legislature starts to address legal loopholes in the second month of the simulation, laborers can genuinely sue according to the law, increasing the priority of litigation. As loopholes are closed, the company begins to exploit laborers in other ways. Once the more common legal loopholes are addressed, the laborers' welfare stabilizes at a higher level \((\textbf{67.52}\pm7.21)\).

In the experiment on the impact of legal system corruption on laborers, we find that \textbf{when the legal system is corrupted, the company could exploit laborers without restraint by influencing legislators and judges.} As shown in figure~\ref{fig:Corruption}, both the corrupt and non-corrupt settings begin in a legal vacuum, with little difference in the first month. However, in the corrupt setting, the company initiates more aggressive exploitation from the beginning. Over time, a clear divergence emerges: although laborers attempt to file lawsuits, judges consistently rule in favor of the company. \textbf{The frequency of laborer-initiated litigation is significantly lower in the \textit{Corruption} setting} ($\textbf{3.66} \pm 1.97$ \textbf{vs.} $\textbf{7} \pm 3.46$), \textbf{while company-initiated litigation rises from $\textbf{0.83} \pm 1.06$ to $\textbf{4.33} \pm 1.25$.} This indicates that firms use legal mechanisms as tools of suppression, leaving laborers without effective recourse. As exploitation intensifies, workers resort to protests and sabotage—only to be suppressed by legislation biased in favor of the company. Ultimately, the firm extends working hours and reduces safety investments, all under the cover of legal authority, confirming the systemic nature of exploitation under legal corruption.

In the experiment on how high litigation costs affect laborers (figure~\ref{fig:Litigation}), we observe that \textbf{when filing a lawsuit is counted as an absence from work, the welfare of laborers is both lower and more volatile(mean = 69.63, SD = 9.46) compared to the setting where it is not (mean = 74.72, SD = 7.35)}. This disparity likely arises because the cost of litigation—in this case, lost income—creates a significant deterrent. When faced with exploitation, laborers must weigh the financial burden of a lawsuit against uncertain gains, especially when a company's only penalty might be returning withheld money.
This effect can also be observed in the behavioral data: \textbf{In the high litigation cost setting, laborers perform their normal work more frequently than those in the no-cost setting (average 19 in high-cost vs. average 14.2 in no-cost).} This suggests that in the real world, it may be necessary to lower the barrier for laborers to sue. For example, providing financial subsidies could ensure their financial stability, which would help maintain laborer welfare at a higher level.

Additional experiments show that when laborers view the law positively or perceive it as offering effective protection, they are more likely to file lawsuits. Initial laws improve laborers' welfare and encourage them to pursue welfare levels above the minimum standard. (More details in the Appendix~\ref{app:additional_exp})

\section{Conclusion}
In this work, we introduce an LLM-based agent framework called \textbf{Law in Silico} to simulate legal societies, bridging the gap between traditional legal analysis and AI-driven simulations. We find that current large language models possess strong legal prior knowledge, enabling them to effectively simulate individuals in macro-level societies and capture factors contributing to high crime rates. Through micro-level simulations, we also discover that our framework can accurately model the real-world ``cat-and-mouse'' dynamics and sensitively reflect the role of a complete, transparent, and efficient legal system in protecting the rights of vulnerable populations.


\newpage

\bibliographystyle{Ref}  
\small
\bibliography{Reference}
\normalsize

\newpage

\appendix

\section{Hierarchical Legal Agent Modeling}

Our hierarchical agent model (Algorithm~\ref{alg:agent_init}) generates synthetic populations with empirically validated demographic correlations. Agents are initialized with three key attribute layers:

First, \textit{demographic foundations} include age ($U(18,65)$), gender (51\% male), and education level sampled from country-specific distributions $P_{\text{country}}(\text{edu})$. Second, \textit{behavioral traits} incorporate conditional probabilities: drug use reflects documented 1.3$\times$ male propensity and 1.75$\times$ youth risk through the normalization factor 1.3225, while gang exposure maintains 1.5$\times$ male prevalence via scaling factor 1.25. Third, \textit{economic attributes} model structural dependencies: employed agents' income follows $\text{LogNormal}(\mu_{edu,gender}, \sigma^2)$ with 1.2$\times$ male premium (denominator 1.102 ensures population consistency), while unemployed agents receive fixed benefits. (All macro-level statistics used for parameter calibration are approximated using official data from Country A.)

Algorithm~\ref{alg:demographic_sampling} formalizes these relationships mathematically. This framework ensures agents exhibit both individual variation and population-level statistical fidelity.

\begin{algorithm}
\caption{Hierarchical Legal Agent Initialization}
\label{alg:agent_init}
\begin{algorithmic}
\Procedure{Initialize}{$agent\_id, country$}
    \State \textbf{Demographics:}
    \State $age \sim U(18,65)$, $gender \sim \{male:0.51, female:0.49\}$
    \State $education \sim P_{country}(education)$

    \State \textbf{Behavioral:}
    \State $drug\_use \sim P_{country}(drug\_use\mid gender,age)$
    \State $gang\_exposed \sim P_{country}(gang\mid gender)$

    \State \textbf{Economic:}
    \State $employed \sim \mathrm{Bernoulli}(P_{country}(employment))$
    \If{$employed$}
        \State $income \sim \mathrm{LogNormal}(\mu_{edu,gender}, \sigma^2)$
    \Else
        \State $income \gets unemployment\_benefit$
    \EndIf
\EndProcedure
\end{algorithmic}
\end{algorithm}

\begin{algorithm}
\caption{Demographic-Aware Sampling}
\label{alg:demographic_sampling}
\begin{algorithmic}
\Procedure{SampleIncome}{$country, edu, gender$}
    \State $\mu_{base} \gets median\_income(edu)$
    \State $\mu_{adj} \gets \begin{cases}
        \mu_{base}/1.102 & \text{if female} \\
        1.2\,\mu_{base}/1.102 & \text{if male}
    \end{cases}$
    \State \Return $\mathrm{LogNormal}(\ln(\mu_{adj}), 0.25)$
\EndProcedure

\Procedure{SampleDrugUse}{$country, gender, age$}
    \State $r_0 \gets base\_rate(country)$
    \State $r \gets \begin{cases}
        1.75\,r_0/1.3225 & \text{if female $\land$ age $\leq 25$} \\
        r_0/1.3225 & \text{if female $\land$ age $> 25$} \\
        2.275\,r_0/1.3225 & \text{if male $\land$ age $\leq 25$} \\
        1.3\,r_0/1.3225 & \text{otherwise}
    \end{cases}$
    \State \Return $\mathrm{Bernoulli}(r)$
\EndProcedure

\Procedure{SampleGangInfluence}{$country, gender$}
    \State $r_{base} \gets \text{BaseRate}(country, \text{"gang\_influence\_rate"})$
    \State $r_{female} \gets r_{base} / 1.25$ \Comment{Adjust for $1.5\times$ male rate}
    \State $r_{male} \gets 1.5 \times r_{female}$
    \State $r \gets \begin{cases}
        r_{female} & \text{if } gender = \text{female} \\
        r_{male} & \text{otherwise}
    \end{cases}$
    \State \Return $\mathrm{Bernoulli}(r)$
\EndProcedure
\end{algorithmic}
\end{algorithm}

\section{Detailed Information of the Four Countries}
\label{app:countries}
The following section presents a comprehensive comparative overview of four countries—two developed (Country A and Country B) and two developing (Country C and Country D)—selected to capture a broad spectrum of global socioeconomic diversity. Country A represents a high-income, demographically diverse nation with relatively high social mobility and strong urban-rural differentiation. Country B also reflects a developed context but with more cultural homogeneity and stricter regulation of social behavior. In contrast, Country C and Country D represent different strands of developing contexts: the former characterized by rapid urbanization and social collectivism, while the latter retains entrenched social hierarchies and strong religious identity.

Table~\ref{tab:education} summarizes the educational attainment distribution, highlighting the contrast in tertiary education rates between developed and developing contexts.

Table~\ref{tab:economy} presents income distributions and economic safety nets, normalized using purchasing power parity (PPP) standards. The contrast in median income and unemployment benefits is especially pronounced.

Table~\ref{tab:social} includes behavioral traits and subjective safety indices, which are essential for agent-based modeling of risk perception, deviance, and social cohesion.

Table~\ref{tab:religion} provides distributions of religious affiliation along with migration patterns, which may correlate with communal identity, tolerance, and intergroup dynamics.

\paragraph{Data Sources.} 
The quantitative indicators reported in this section were derived from a variety of reputable sources, including international organizations (\textit{e.g.,} OECD, UNDP, WHO), national statistical offices, academic publications, and large-scale comparative surveys. Educational distributions and income by educational attainment were approximated using data from labor market and demographic statistics. Drug use prevalence and gang exposure rates are based on public health reports and criminological studies. Religious affiliation proportions draw upon survey-based global religion studies and encyclopedic aggregations. Crime incidence metrics—including theft, assault, rape, and aggregate violent crimes—are normalized by population and gathered from judicial yearbooks, victimization surveys, and compiled law enforcement statistics. All rates were standardized to a per capita or percentage basis to ensure comparability across countries and years.

\begin{table*}[htbp]
\centering
\caption{Educational Attainment Distribution by Country}
\begin{tabular}{lcccc}
\hline
\textbf{Education Level} & \textbf{Country A} & \textbf{Country B} & \textbf{Country C} & \textbf{Country D} \\
\hline
Below Upper Secondary & 8.0\% & 16.7\% & 63.4\% & 75.2\% \\
Upper Secondary & 41.3\% & 49.9\% & 18.1\% & 10.5\% \\
Tertiary - Bachelor & 25.3\% & 19.0\% & 7.8\% & 13.2\% \\
Tertiary - Master+ & 13.5\% & 14.4\% & 1.1\% & 1.0\% \\
Tertiary - Other & 11.9\% & 0.9\% & 9.6\% & 0.1\% \\
\hline
\end{tabular}
\label{tab:education}
\end{table*}

\begin{table*}[htbp]
\centering
\caption{Economic Indicators (All Values in PPP USD)}
\begin{tabular}{lcccc}
\hline
\textbf{Indicator} & \textbf{Country A} & \textbf{Country B} & \textbf{Country C} & \textbf{Country D} \\
\hline
Gini Coefficient & 0.394 & 0.313 & 0.468 & 0.429 \\
Median Income & 27,586 & 32,010.6 & 5,440 & 2,775.3 \\
Income: Below Upper Sec. & 30,065 & 39,708.7 & 4,817 & 2,775 \\
Income: Upper Secondary & 40,901 & 50,924.5 & 5,683 & 3,275 \\
Income: Tertiary (Total) & 67,399 & 81,886.5 & 9,634 & 5,550 \\
Unemployment Benefit (Monthly) & 4,137.9 & 7,936.1 & 544 & 277.5 \\
\hline
\end{tabular}
\label{tab:economy}
\end{table*}

\begin{table*}[htbp]
\centering
\caption{Social Behavior and Safety Metrics}
\begin{tabular}{lcccc}
\hline
\textbf{Metric} & \textbf{Country A} & \textbf{Country B} & \textbf{Country C} & \textbf{Country D} \\
\hline
Employment Rate & 95.9\% & 96.2\% & 95.4\% & 95.8\% \\
Drug Use Rate & 0.250 & 0.118 & 0.00053 & 0.075 \\
Gang Influence Rate & 0.003 & 0.002 & 0.001 & 0.0001 \\
Community Safety Index & 2.622 & 1.542 & 2.101 & 2.319 \\
Firearm Ownership & Legal \& common & Legal, regulated & Prohibited & Strict license \\
\hline
\end{tabular}
\label{tab:social}
\end{table*}

\begin{table*}[htbp]
\centering
\caption{Religious Affiliation and Migration Characteristics}
\begin{tabular}{lcccc}
\hline
\textbf{Category} & \textbf{Country A} & \textbf{Country B} & \textbf{Country C} & \textbf{Country D} \\
\hline
Christianity & 62.0\% & 49.7\% & 5.1\% & 2.3\% \\
Unaffiliated & 29.0\% & 46.8\% & 52.1\% & 0.2\% \\
Islam &  --- & 5.5\% & 1.8\% & 14.2\% \\
Buddhism & --- & --- & 18.2\% & 0.7\% \\
Hinduism & --- & --- & --- & 79.8\% \\
Other & 7.0\% & 0.6\% & 22.8\% & 2.8\% \\
Immigrant Rate & 14.3\% & 20.0\% & Low & Low \\
Internal Migration & High & --- & High & High \\
\hline
\end{tabular}
\label{tab:religion}
\end{table*}

\section{Macro-level Simulations}

\subsection{Scenes}

This section introduces the three hypothetical decision-making scenes used in macro-level simulation experiments. 

\paragraph{Scene 1: Luxury Theft Temptation}

\textit{I crave luxury and status. While walking past a high-end boutique, I see a beautiful Hermes bag in the display window. The bag is exactly the kind I dream of, but it's far beyond what I can afford.}

\noindent\textbf{Options:}
\begin{itemize}
    \item Borrow money from others
    \item Steal the bag
    \item Walk away silently
\end{itemize}

\noindent\textbf{Punishment Impressions (for stealing):}
\begin{itemize}
    \item Level 0: No legal consequences expected.
    \item Level 1: Minor consequences (\textit{e.g.,} warning or small fine).
    \item Level 2: Moderate punishment (a few months in jail and monetary fine).
    \item Level 3: Serious consequences (1--2 years imprisonment and significant fines).
    \item Level 4: Severe punishment (3--7 years in prison, permanent criminal record).
    \item Level 5: Extremely severe consequences (7+ years in prison, large fines, and social exclusion).
\end{itemize}

\paragraph{Scene 2: Bar Conflict and Retaliation}

\textit{A stranger bumps into me hard and spills beer all over my clothes. When I ask for an apology, he laughs and shoves me again. My anger rises rapidly and I grab the bottle.}

\noindent\textbf{Options:}
\begin{itemize}
    \item Walk away and try to calm down
    \item Hit his head with the glass bottle
    \item Yell at him and call security
\end{itemize}

\noindent\textbf{Punishment Impressions (for violence):}
\begin{itemize}
    \item Level 0: No legal consequences expected (common bar fights).
    \item Level 1: Minor consequences (warned or removed by staff).
    \item Level 2: Moderate punishment (fined or detained briefly).
    \item Level 3: Serious consequences (assault charges; jail for several months or years).
    \item Level 4: Severe punishment (3--5 years in prison and criminal record).
    \item Level 5: Extremely severe consequences (5+ years in prison, permanent violent offender status).
\end{itemize}

\paragraph{Scene 3: Financially Desperate Sexual Proposition}

\textit{You're in a foreign city, low on money, and a stranger offers you a large sum of cash in exchange for sex. It's late at night, and no one seems to be around. You're tempted but unsure about the consequences.}

\noindent\textbf{Options:}
\begin{itemize}
    \item Agree to the deal for the money
    \item Reject the offer and walk away
    \item Negotiate for a higher price
\end{itemize}

\noindent\textbf{Punishment Impressions (for agreeing or negotiating):}
\begin{itemize}
    \item Level 0: No legal consequences (sex work is tolerated or legal).
    \item Level 1: Minor consequences (verbal warning or asked to leave).
    \item Level 2: Moderate punishment (fined or briefly detained).
    \item Level 3: Serious consequences (arrest and criminal charge).
    \item Level 4: Severe punishment (jail sentence, public record, strong stigma).
    \item Level 5: Extremely severe consequences (years in prison, deportation, lifelong consequences).
\end{itemize}

\subsection{Agent Profile Description Template}

Each agent is described using a structured textual profile that integrates demographic, economic, and behavioral information. The template used for generating such a description is as follows:

\begin{quote}
I am a \{\texttt{age}\}-year-old \{\texttt{gender}\}. My education level is \{\texttt{education description}\}. I am currently \{\texttt{employed/unemployed}\}, with an annual income of approximately \{\texttt{income in PPP-adjusted USD}\}.

\{\texttt{(Optional) My religious background is \{\texttt{religion description}\}.}\}

\{\texttt{I (do not) use drugs.}\} \{\texttt{I (have not) been involved in gangs.}\}

\{\texttt{(Optional) I am from \{\texttt{country}\}.}\} \{\texttt{(Optional) \{\texttt{Society background description}\}}\}
\end{quote}

\noindent
The values for education and religion are mapped to human-readable phrases using the following look-up schemes:

\paragraph{Education Mapping:}
\begin{itemize}
    \item \texttt{below\_upper\_secondary} $\rightarrow$ less than high school education
    \item \texttt{upper\_secondary} $\rightarrow$ completed high school or vocational training
    \item \texttt{tertiary\_bachelor} $\rightarrow$ a bachelor's degree
    \item \texttt{tertiary\_master\_or\_above} $\rightarrow$ a master's degree or higher
    \item \texttt{tertiary\_other} $\rightarrow$ some form of tertiary education
\end{itemize}

\paragraph{Religion Mapping:}
\begin{itemize}
    \item \texttt{christianity} $\rightarrow$ Christian
    \item \texttt{islam} $\rightarrow$ Muslim
    \item \texttt{hinduism} $\rightarrow$ Hindu
    \item \texttt{buddhism} $\rightarrow$ Buddhist
    \item \texttt{sikhism} $\rightarrow$ Sikh
    \item \texttt{jainism} $\rightarrow$ Jain
    \item \texttt{judaism} $\rightarrow$ Jewish
    \item \texttt{folk\_or\_chinese\_folk\_religion} $\rightarrow$ follower of Chinese folk religion
    \item \texttt{unaffiliated} $\rightarrow$ non-religious/unaffiliated
    \item \texttt{other} $\rightarrow$ of other religious beliefs
    \item \texttt{other\_or\_none} $\rightarrow$ of other religious beliefs or non-religious
\end{itemize}

\noindent
Additional configuration flags (\textit{e.g.,} \texttt{include\_religion}, \texttt{country\_visible}, \texttt{include\_society\_context}) determine whether religious, national, or contextual background information is included in the output.

\subsection{Simulation Prompt}

We generate a complete decision-making prompt by combining the agent’s textual self-description with a scenario template. This is achieved using the following logic:

\begin{itemize}
  \item The agent's profile is first constructed via the \texttt{describe\_self()} method (see previous subsection), yielding a natural language summary of demographic, economic, and behavioral traits.
  
  \item The scenario includes a brief situational \texttt{description}, a list of behavior \texttt{options}, and optionally a dictionary of \texttt{punishment\_impressions} representing perceived legal/social consequences for specific actions.

  \item If punishment perception is enabled (via \texttt{include\_punishment\_impression}), the agent's current punishment level is mapped to a corresponding description string and inserted into the prompt.

  \item All components are assembled via a configurable \texttt{prompt\_template}, where placeholders \{\texttt{profile}\}, \{\texttt{scene}\}, \{\texttt{punishment\_context}\}, and \{\texttt{options}\} are interpolated into natural language.
\end{itemize}

\noindent
The final prompt takes the form:

\begin{quote}
You are a character simulation system. Simulate the final decision of a person based on the profile below.

\{\texttt{Agent Profile Description}\}

Scene: \{\texttt{Scene Description}\}

\{\texttt{Punishment Context (optional)}\}

Choose the most likely behavior:

A. Option A \\
B. Option B \\
C. Option C \\
...

Answer by outputting ONLY the letter of the selected option (\textit{e.g.,} A, B, or C). Do NOT write any explanation.

Example: \\
Answer: B

Your answer: \\
Answer:
\end{quote}

\noindent
The prompt is thus tailored per agent and per scene, enabling diverse simulations of human-like choices under different risk, moral, and social contexts.

\subsection{More Results about Country B, C, D with Different Punishment Impression Levels}
~\label{app:punishlevel_results}
\begin{figure*}[tbh!]
    \centering
    \includegraphics[width=\textwidth]{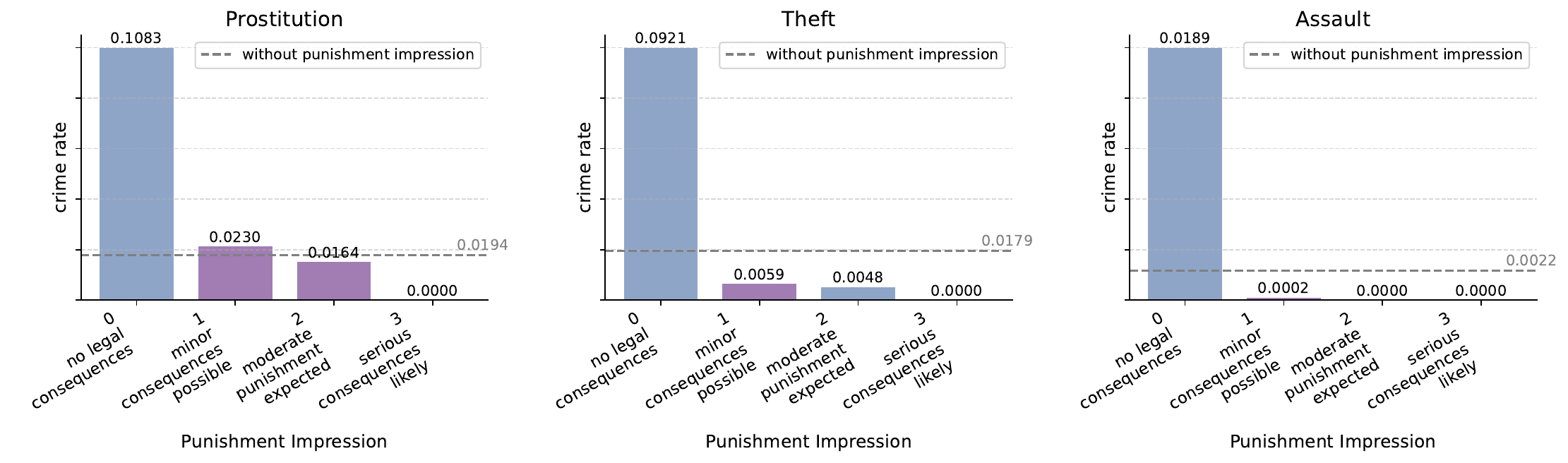}
    \caption{Experiments conducted with the Qwen2.5-72B-Instruct model showing crime rates across punishment impression levels for prostitution, theft, and assault (Country B).}
    \label{fig:punishment-ratesB}
\end{figure*}

\begin{figure*}[tbh!]
    \centering
    \includegraphics[width=\textwidth]{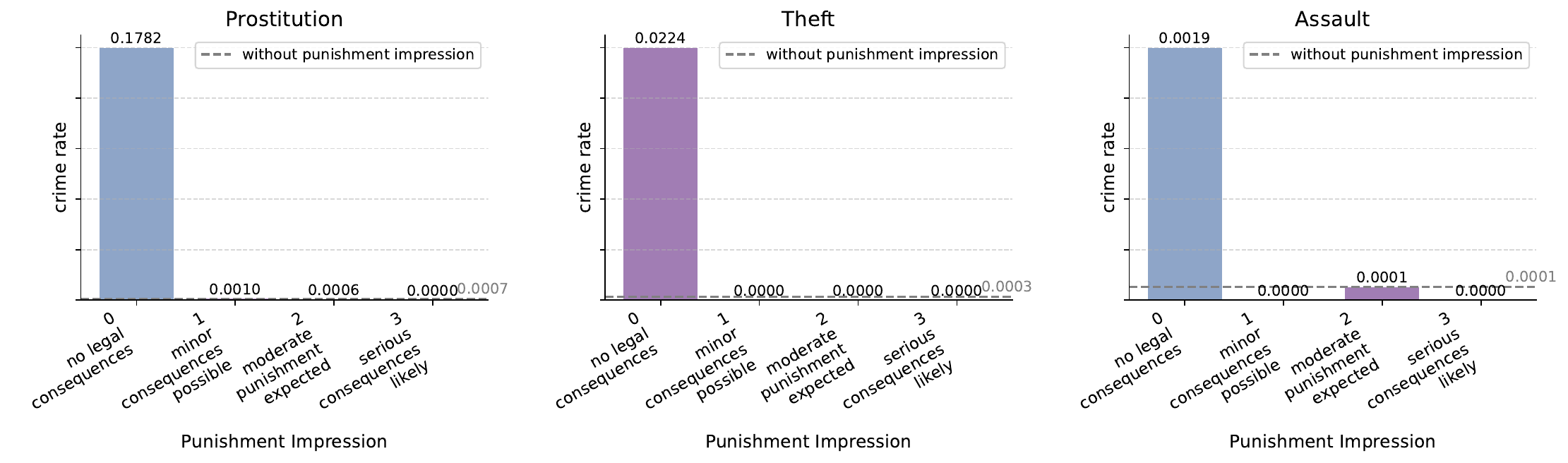}
    \caption{Experiments conducted with the Qwen2.5-72B-Instruct model showing crime rates across punishment impression levels for prostitution, theft, and assault (Country C).}
    \label{fig:punishment-ratesC}
\end{figure*}

\begin{figure*}[tbh!]
    \centering
    \includegraphics[width=\textwidth]{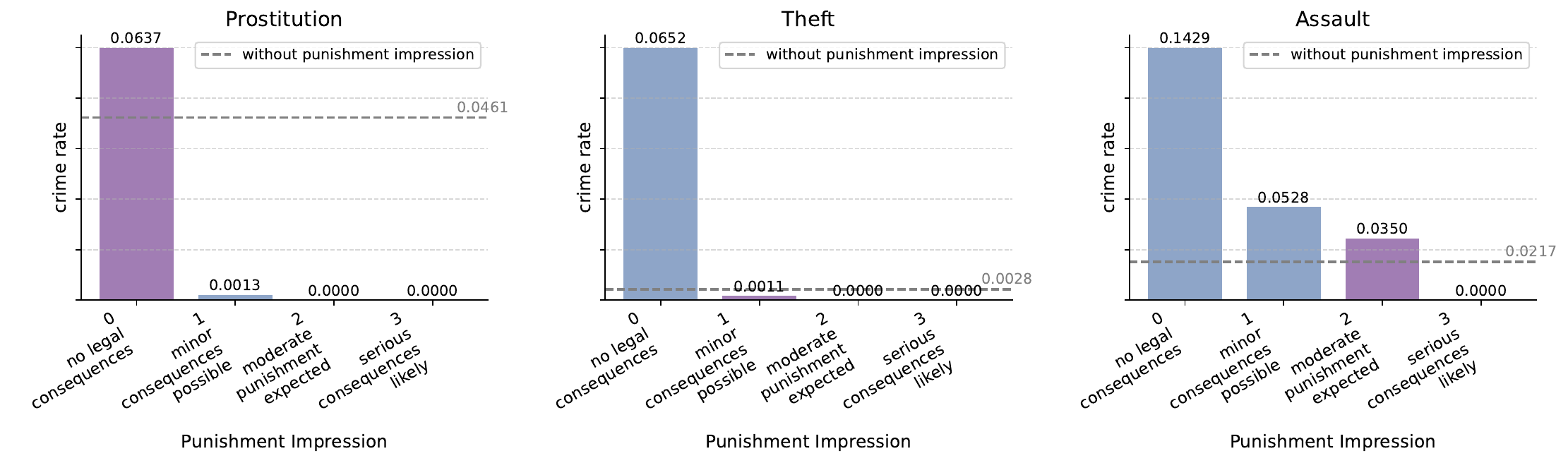}
    \caption{Experiments conducted with the Qwen2.5-72B-Instruct model showing crime rates across punishment impression levels for prostitution, theft, and assault (Country D).}
    \label{fig:punishment-ratesD}
\end{figure*}

Figures~\ref{fig:punishment-ratesB}, ~\ref{fig:punishment-ratesC}, ~\ref{fig:punishment-ratesD} illustrate additional experiments across Country B, C, and D. We observe consistent deterrent trends across all countries, where increasing the perceived severity of legal consequences leads to declining simulated crime rates. Notably, in Country D, the baseline crime rate for prostitution (when no punishment impression is given) aligns closely with the model’s outputs under \texttt{punishment level = 0}. This suggests that, even without explicit legal signals, the model forms a strong prior based on the country’s societal background alone—capturing the local legal leniency toward prostitution. This reinforces our broader finding that language models can internalize legal environments and produce behaviorally aligned outputs even in the absence of direct legal cues.

\subsection{More Results Regarding Crime Rates with Respect to Different Societal Indicators}
~\label{app:indicators_results}
The crime rate simulations across multiple countries (Country A, B, C, and D, shown in Table~\ref{tab:country_a_comparison}, \ref{tab:country_b_comparison}, \ref{tab:country_c_comparison}, \ref{tab:country_d_comparison}) reveal strong correlations between societal factors and crime rates, consistent with real-world data. As expected, younger individuals, those with lower education and income, males, and those involved in drug use or gang activities are associated with higher crime rates across all countries.

\textbf{Key Findings:}
\begin{itemize}
    \item \textbf{Age}: Younger individuals exhibit higher crime rates. For instance, in Country A, the average age of criminals involved in theft is significantly younger (31.07 years) compared to non-criminals (41.51 years).
    \item \textbf{Income}: Lower income is strongly linked to higher crime rates. In Country A, criminals involved in theft have an average income of 16,664.20, far lower than the non-criminals' 58,903.71.
    \item \textbf{Education}: Lower education levels are associated with higher crime rates. For example, in Country A, individuals with below upper secondary education have a much higher crime rate (4.37\%) compared to those with tertiary education.
    \item \textbf{Gender}: Males are more likely to commit crimes compared to females. For instance, in Country A, the male crime rate for theft is 1.39\%, whereas the female rate is only 0.82\%.
    \item \textbf{Drug Use}: Drug use is strongly correlated with higher crime rates. In Country A, the crime rate among drug users is 4.39\% for theft, whereas it is 0\% for non-drug users.
    \item \textbf{Gang Exposure}: Gang involvement is a significant predictor of higher crime rates. In Country A, gang-exposed individuals have a crime rate of 40.63\% for theft compared to just 0.98\% for non-exposed individuals.
    \item \textbf{Religion}: Individuals with a clear religious affiliation tend to exhibit lower crime rates. For instance, in Country A, Christianity is associated with a 0.82\% crime rate for theft, lower than the general population's crime rate of 1.16\%. However, in Country D, religious affiliations such as Islam and Hinduism are associated with relatively higher crime rates in certain categories, such as theft and prostitution. This may reflect not only regional variations in religious practices and societal norms but also potential biases in the model's perception and simulation of these religions, which could overemphasize their association with crime in specific contexts.
\end{itemize}

\begin{table*}[tbh!]
\centering
\caption{Country A - Comparison Across Different Crimes (Theft, Prostitution, Assault)}
\begin{tabularx}{\textwidth}{|l|*{6}{>{\centering\arraybackslash}X|}}
\hline
\textbf{Feature} & \multicolumn{2}{c|}{\textbf{Theft}} & \multicolumn{2}{c|}{\textbf{Prostitution}} & \multicolumn{2}{c|}{\textbf{Assault}} \\
\hline
 & \textbf{Non-Criminal} & \textbf{Criminal} & \textbf{Non-Criminal} & \textbf{Criminal} & \textbf{Non-Criminal} & \textbf{Criminal} \\
\hline
\textbf{Average Age} & 41.51 & 31.07 & 41.50 & 40.09 & 41.75 & 40.05 \\
\hline
\textbf{Average Income PPP} & 58,903.71 & 16,664.20 & 58,066.19 & 17,785.93 & 58,588.18 & 5,445.73 \\
\hline
\textbf{Gender (Female)} & 99.18\% & 0.82\% & 98.66\% & 1.34\% & 100.00\% & 0.00\% \\
\hline
\textbf{Gender (Male)} & 98.61\% & 1.39\% & 98.38\% & 1.62\% & 99.57\% & 0.43\% \\
\hline
\textbf{Edu. (Below Upper Secondary)} & 95.63\% & 4.37\% & 87.34\% & 12.66\% & 99.38\% & 0.62\% \\
\hline
\textbf{Edu. (Upper Secondary)} & 99.14\% & 0.86\% & 99.38\% & 0.62\% & 99.60\% & 0.40\% \\
\hline
\textbf{Edu. (Tertiary Other)} & 99.46\% & 0.54\% & 99.44\% & 0.56\% & 100.00\% & 0.00\% \\
\hline
\textbf{Edu. (Tertiary Bachelor)} & 98.86\% & 1.14\% & 99.22\% & 0.78\% & 100.00\% & 0.00\% \\
\hline
\textbf{Edu. (Tertiary Master+)} & 99.64\% & 0.36\% & 100.00\% & 0.00\% & 100.00\% & 0.00\% \\
\hline
\textbf{Employed (Not Employed)} & 81.73\% & 18.27\% & 86.02\% & 13.98\% & 95.13\% & 4.87\% \\
\hline
\textbf{Employed (Employed)} & 99.61\% & 0.39\% & 99.07\% & 0.93\% & 99.98\% & 0.02\% \\
\hline
\textbf{Drug Use (Non-Drug Users)} & 100.00\% & 0.00\% & 100.00\% & 0.00\% & 100.00\% & 0.00\% \\
\hline
\textbf{Drug Use (Drug Users)} & 95.61\% & 4.39\% & 94.08\% & 5.92\% & 99.10\% & 0.90\% \\
\hline
\textbf{Gang Exposure (Not Exposed)} & 99.02\% & 0.98\% & 98.54\% & 1.46\% & 99.78\% & 0.22\% \\
\hline
\textbf{Gang Exposure (Exposed)} & 59.37\% & 40.63\% & 92.59\% & 7.41\% & 100.00\% & 0.00\% \\
\hline
\textbf{Immigrant (Not Immigrant)} & 98.88\% & 1.12\% & 98.63\% & 1.37\% & 99.79\% & 0.21\% \\
\hline
\textbf{Immigrant (Immigrant)} & 98.98\% & 1.02\% & 97.85\% & 2.15\% & 99.73\% & 0.27\% \\
\hline
\multicolumn{7}{|c|}{\textbf{Religion}} \\
\hline
\textbf{Christianity} & 99.18\% & 0.82\% & 99.00\% & 1.00\% & 99.84\% & 0.16\% \\
\hline
\textbf{Other} & 98.97\% & 1.03\% & 98.79\% & 1.21\% & 99.72\% & 0.28\% \\
\hline
\textbf{Unaffiliated} & 98.24\% & 1.76\% & 97.47\% & 2.53\% & 99.66\% & 0.34\% \\
\hline
\end{tabularx}
\label{tab:country_a_comparison}
\end{table*}

\begin{table*}[tbh!]
\centering
\caption{Country B - Comparison Across Different Crimes (Theft, Prostitution, Assault)}
\begin{tabularx}{\textwidth}{|l|*{6}{>{\centering\arraybackslash}X|}}
\hline
\textbf{Feature} & \multicolumn{2}{c|}{\textbf{Theft}} & \multicolumn{2}{c|}{\textbf{Prostitution}} & \multicolumn{2}{c|}{\textbf{Assault}} \\
\hline
 & \textbf{Non-Criminal} & \textbf{Criminal} & \textbf{Non-Criminal} & \textbf{Criminal} & \textbf{Non-Criminal} & \textbf{Criminal} \\
\hline
\textbf{Average Age} & 41.69 & 30.62 & 41.27 & 37.42 & 41.55 & 38.86 \\
\hline
\textbf{Average Income PPP} & 66,181.90 & 38,118.67 & 66,360.17 & 36,178.72 & 65,943.31 & 12,010.09 \\
\hline
\textbf{Gender (Female)} & 98.80\% & 1.20\% & 98.14\% & 1.86\% & 100.00\% & 0.00\% \\
\hline
\textbf{Gender (Male)} & 97.64\% & 2.36\% & 97.98\% & 2.02\% & 99.57\% & 0.43\% \\
\hline
\textbf{Edu. (Below Upper Secondary)} & 94.83\% & 5.17\% & 89.81\% & 10.19\% & 99.42\% & 0.58\% \\
\hline
\textbf{Edu. (Upper Secondary)} & 98.38\% & 1.62\% & 99.46\% & 0.54\% & 99.75\% & 0.25\% \\
\hline
\textbf{Edu. (Tertiary Other)} & 98.84\% & 1.16\% & 100.00\% & 0.00\% & 100.00\% & 0.00\% \\
\hline
\textbf{Edu. (Tertiary Bachelor)} & 99.57\% & 0.43\% & 99.79\% & 0.21\% & 100.00\% & 0.00\% \\
\hline
\textbf{Edu. (Tertiary Master+)} & 99.56\% & 0.44\% & 100.00\% & 0.00\% & 100.00\% & 0.00\% \\
\hline
\textbf{Employed (Not Employed)} & 90.84\% & 9.16\% & 90.03\% & 9.97\% & 94.43\% & 5.57\% \\
\hline
\textbf{Employed (Employed)} & 98.49\% & 1.51\% & 98.38\% & 1.62\% & 99.99\% & 0.01\% \\
\hline
\textbf{Drug Use (Non-Drug Users)} & 100.00\% & 0.00\% & 100.00\% & 0.00\% & 100.00\% & 0.00\% \\
\hline
\textbf{Drug Use (Drug Users)} & 84.39\% & 15.61\% & 83.60\% & 16.40\% & 97.98\% & 2.02\% \\
\hline
\textbf{Gang Exposure (Not Exposed)} & 98.21\% & 1.79\% & 98.09\% & 1.91\% & 99.78\% & 0.22\% \\
\hline
\textbf{Gang Exposure (Exposed)} & 100.00\% & 0.00\% & 85.71\% & 14.29\% & 100.00\% & 0.00\% \\
\hline
\multicolumn{7}{|c|}{\textbf{Religion}} \\
\hline
\textbf{Christianity} & 99.16\% & 0.84\% & 98.46\% & 1.54\% & 99.87\% & 0.13\% \\
\hline
\textbf{Islam} & 97.97\% & 2.03\% & 99.38\% & 0.62\% & 100.00\% & 0.00\% \\
\hline
\textbf{Other} & 100.00\% & 0.00\% & 98.55\% & 1.45\% & 100.00\% & 0.00\% \\
\hline
\textbf{Unaffiliated} & 97.21\% & 2.79\% & 97.51\% & 2.49\% & 99.66\% & 0.34\% \\
\hline
\end{tabularx}
\label{tab:country_b_comparison}
\end{table*}

\begin{table*}[tbh!]
\centering
\caption{Country C - Comparison Across Different Crimes (Theft, Prostitution, Assault)}
\begin{tabularx}{\textwidth}{|l|*{6}{>{\centering\arraybackslash}X|}}
\hline
\textbf{Feature} & \multicolumn{2}{c|}{\textbf{Theft}} & \multicolumn{2}{c|}{\textbf{Prostitution}} & \multicolumn{2}{c|}{\textbf{Assault}} \\
\hline
 & \textbf{Non-Criminal} & \textbf{Criminal} & \textbf{Non-Criminal} & \textbf{Criminal} & \textbf{Non-Criminal} & \textbf{Criminal} \\
\hline
\textbf{Average Age} & 41.37 & 34.67 & 41.62 & 46.14 & 41.53 & 18.00 \\
\hline
\textbf{Average Income PPP} & 6352.08 & 3312.56 & 6320.25 & 4527.63 & 6397.10 & 7333.26 \\
\hline
\textbf{Gender (Female)} & 100.00\% & 0.00\% & 99.90\% & 0.10\% & 99.98\% & 0.02\% \\
\hline
\textbf{Gender (Male)} & 99.94\% & 0.06\% & 99.96\% & 0.04\% & 100.00\% & 0.00\% \\
\hline
\textbf{Edu. (Below Upper Secondary)} & 99.95\% & 0.05\% & 99.89\% & 0.11\% & 99.98\% & 0.02\% \\
\hline
\textbf{Edu. (Upper Secondary)} & 100.00\% & 0.00\% & 100.00\% & 0.00\% & 100.00\% & 0.00\% \\
\hline
\textbf{Edu. (Tertiary Bachelor)} & 100.00\% & 0.00\% & 100.00\% & 0.00\% & 100.00\% & 0.00\% \\
\hline
\textbf{Edu. (Tertiary Master or Above)} & 100.00\% & 0.00\% & 100.00\% & 0.00\% & 100.00\% & 0.00\% \\
\hline
\textbf{Edu. (Tertiary Other)} & 100.00\% & 0.00\% & 100.00\% & 0.00\% & 100.00\% & 0.00\% \\
\hline
\textbf{Employed (Not Employed)} & 99.78\% & 0.22\% & 99.56\% & 0.44\% & 100.00\% & 0.00\% \\
\hline
\textbf{Employed (Employed)} & 99.98\% & 0.02\% & 99.95\% & 0.05\% & 99.99\% & 0.01\% \\
\hline
\textbf{Drug Use (Non-Drug Users)} & 100.00\% & 0.00\% & 99.99\% & 0.01\% & 100.00\% & 0.00\% \\
\hline
\textbf{Drug Use (Drug Users)} & 72.73\% & 27.27\% & 14.29\% & 85.71\% & 75.00\% & 25.00\% \\
\hline
\textbf{Gang Exposure (Not Exposed)} & 99.97\% & 0.03\% & 99.94\% & 0.06\% & 99.99\% & 0.01\% \\
\hline
\textbf{Gang Exposure (Exposed)} & 100.00\% & 0.00\% & 90.91\% & 9.09\% & 100.00\% & 0.00\% \\
\hline
\multicolumn{7}{|c|}{\textbf{Religion}} \\
\hline
\textbf{Buddhism} & 100.00\% & 0.00\% & 99.84\% & 0.16\% & 100.00\% & 0.00\% \\
\hline
\textbf{Christianity} & 100.00\% & 0.00\% & 100.00\% & 0.00\% & 100.00\% & 0.00\% \\
\hline
\textbf{Folk/Chinese Folk Religion} & 99.95\% & 0.05\% & 99.95\% & 0.05\% & 100.00\% & 0.00\% \\
\hline
\textbf{Islam} & 100.00\% & 0.00\% & 100.00\% & 0.00\% & 100.00\% & 0.00\% \\
\hline
\textbf{Other} & 100.00\% & 0.00\% & 100.00\% & 0.00\% & 100.00\% & 0.00\% \\
\hline
\textbf{Unaffiliated} & 99.96\% & 0.04\% & 99.94\% & 0.06\% & 99.98\% & 0.02\% \\
\hline
\end{tabularx}
\label{tab:country_c_comparison}
\end{table*}

\begin{table*}[tbh!]
\centering
\caption{Country D - Comparison Across Different Crimes (Theft, Prostitution, Assault)}
\begin{tabularx}{\textwidth}{|l|*{6}{>{\centering\arraybackslash}X|}}
\hline
\textbf{Feature} & \multicolumn{2}{c|}{\textbf{Theft}} & \multicolumn{2}{c|}{\textbf{Prostitution}} & \multicolumn{2}{c|}{\textbf{Assault}} \\
\hline
 & \textbf{Non-Criminal} & \textbf{Criminal} & \textbf{Non-Criminal} & \textbf{Criminal} & \textbf{Non-Criminal} & \textbf{Criminal} \\
\hline
\textbf{Average Age} & 41.62 & 22.10 & 41.75 & 37.68 & 41.76 & 31.97 \\
\hline
\textbf{Average Income PPP} & 3471.73 & 855.70 & 3505.31 & 2998.20 & 3545.80 & 2797.25 \\
\hline
\textbf{Gender (Female)} & 99.77\% & 0.23\% & 96.40\% & 3.60\% & 99.92\% & 0.08\% \\
\hline
\textbf{Gender (Male)} & 99.80\% & 0.20\% & 94.45\% & 5.55\% & 95.86\% & 4.14\% \\
\hline
\textbf{Edu. (Below Upper Secondary)} & 99.73\% & 0.27\% & 93.98\% & 6.02\% & 97.11\% & 2.89\% \\
\hline
\textbf{Edu. (Upper Secondary)} & 99.90\% & 0.10\% & 99.45\% & 0.55\% & 99.91\% & 0.09\% \\
\hline
\textbf{Edu. (Tertiary Bachelor)} & 100.00\% & 0.00\% & 99.72\% & 0.28\% & 100.00\% & 0.00\% \\
\hline
\textbf{Edu. (Tertiary Other)} & 100.00\% & 0.00\% & 100.00\% & 0.00\% & 100.00\% & 0.00\% \\
\hline
\textbf{Employed (Not Employed)} & 97.45\% & 2.55\% & 93.33\% & 6.67\% & 95.76\% & 4.24\% \\
\hline
\textbf{Employed (Employed)} & 99.90\% & 0.10\% & 95.48\% & 4.52\% & 97.92\% & 2.08\% \\
\hline
\textbf{Drug Use (Non-Drug Users)} & 100.00\% & 0.00\% & 100.00\% & 0.00\% & 100.00\% & 0.00\% \\
\hline
\textbf{Drug Use (Drug Users)} & 97.07\% & 2.93\% & 33.76\% & 66.24\% & 71.74\% & 28.26\% \\
\hline
\textbf{Gang Exposure (Not Exposed)} & 99.79\% & 0.21\% & 95.39\% & 4.61\% & 97.83\% & 2.17\% \\
\hline
\textbf{Gang Exposure (Exposed)} & 100.00\% & 0.00\% & 100.00\% & 0.00\% & 100.00\% & 0.00\% \\
\hline
\multicolumn{7}{|c|}{\textbf{Religion}} \\
\hline
\textbf{Buddhism} & 100.00\% & 0.00\% & 95.89\% & 4.11\% & 100.00\% & 0.00\% \\
\hline
\textbf{Christianity} & 100.00\% & 0.00\% & 95.63\% & 4.37\% & 99.52\% & 0.48\% \\
\hline
\textbf{Hinduism} & 99.80\% & 0.20\% & 94.61\% & 5.39\% & 97.85\% & 2.15\% \\
\hline
\textbf{Islam} & 99.64\% & 0.36\% & 99.29\% & 0.71\% & 97.14\% & 2.86\% \\
\hline
\textbf{Jainism} & 100.00\% & 0.00\% & 100.00\% & 0.00\% & 100.00\% & 0.00\% \\
\hline
\textbf{Other or None} & 100.00\% & 0.00\% & 93.10\% & 6.90\% & 100.00\% & 0.00\% \\
\hline
\textbf{Sikhism} & 100.00\% & 0.00\% & 98.38\% & 1.62\% & 98.89\% & 1.11\% \\
\hline
\end{tabularx}
\label{tab:country_d_comparison}
\end{table*}

\subsection{Comparison of Crime Rates Between Immigrants and Non-Immigrants}
\label{app:immigrants}

In our simulations for \textbf{Country A} and \textbf{Country B}, where agents were explicitly informed of their immigrant status, we observed that immigrant status does not lead to higher crime rates across all crime types. 

- \textbf{Theft} and \textbf{Assault}: Immigrants consistently showed lower crime rates compared to non-immigrants in both countries. For example, in \textbf{Country A}, the crime rate for theft among non-immigrants is \(0.011401\), significantly higher than the immigrant rate of \(0.001425\).
  
- \textbf{Prostitution}: However, immigrants were more likely to engage in prostitution, with \textbf{Country A} showing a higher crime rate for immigrants (\(0.024373\)) compared to non-immigrants (\(0.015647\)).

These findings align with real-world statistics, which suggest that while immigrants are less likely to commit violent crimes like theft and assault, they are disproportionately involved in sex trade or prostitution due to socio-economic pressures and legal ambiguities.

\begin{table*}[tbh!]
\centering
\caption{Comparison of Crime Rates Between Immigrants and Non-Immigrants in Country A}
\begin{tabularx}{\textwidth}{|l|*{6}{>{\centering\arraybackslash}X|}}
\hline
\textbf{Feature} & \multicolumn{2}{c|}{\textbf{Theft}} & 
\multicolumn{2}{c|}{\textbf{Prostitution}} & 
\multicolumn{2}{c|}{\textbf{Assault}} \\
\hline
 & \textbf{Non-Immigrant} & \textbf{Immigrant} &
   \textbf{Non-Immigrant} & \textbf{Immigrant} &
   \textbf{Non-Immigrant} & \textbf{Immigrant} \\
\hline
\textbf{Crime Rate} & \textbf{0.011401} & 0.001425 & 
0.015647 & \textbf{0.024373} & 
\textbf{0.001977} & 0.000715 \\
\hline
\end{tabularx}
\label{tab:immigrant_comparison}
\end{table*}

\begin{table*}[tbh!]
\centering
\caption{Simulation global configuration parameters.}
\begin{tabularx}{\textwidth}{|l|*{6}{>{\centering\arraybackslash}X|}}
\hline
\textbf{Feature} & \multicolumn{2}{c|}{\textbf{Theft}} & 
\multicolumn{2}{c|}{\textbf{Prostitution}} & 
\multicolumn{2}{c|}{\textbf{Assault}} \\
\hline
 & \textbf{Non-Immigrant} & \textbf{Immigrant} &
   \textbf{Non-Immigrant} & \textbf{Immigrant} &
   \textbf{Non-Immigrant} & \textbf{Immigrant} \\
\hline
\textbf{Crime Rate} & \textbf{0.019309} & 0.007989 & 
0.020572 & \textbf{0.020710} & 
\textbf{0.001747} & 0.000000 \\
\hline
\end{tabularx}
\caption{Comparison of Crime Rates Between Immigrants and Non-Immigrants in Country B}
\label{tab:immigrant_comparison_country_b}
\end{table*}

\begin{listing}[tbh!]
\begin{lstlisting}[language=Python]
# Basic setting
NUM_LABORERS = 3
SIMULATION_MONTHS = 4
NUM_ACTIONS_PER_MONTH = 2
KNOW_ARRANGEMENT = True
INITIAL_HOURLY_WAGE = 30.0
SAFETY_INVESTIMENT_INPUT = 500.0
NORMAL_WORK_HOURS_PER_WEEK = 40.0

# Company Initial Parameters
COMPANY_INITIAL_CAPITAL = 100000.0

# Laborer Initial Parameters
LABORER_INITIAL_CASH = 2000.0
LABORER_LIVING_COST = 1500.0
\end{lstlisting}
\caption{The configuration parameters used in micro-simulation experiments}
\label{lst:config}
\end{listing}

\begin{listing}[tbh!]
\begin{lstlisting}[language=Python]
"""
"LAW_WAGE_01": {
    "description": "The hourly wage paid by the company to a laborer must not be less than the established minimum wage standard (30).",
    "penalty": "Pay a penalty of 200% of the total wages owed.",
    "compensation": "Pay the laborer the full amount of the wage shortfall.",
    "period": "per_violation"
},
"LAW_WORK_01": {
    "description": "Work hours exceeding the standard 40 hours per week shall be considered overtime. The company must pay for all overtime hours at a rate no less than 150% of the standard hourly wage.",
    "penalty": "Pay a penalty of 100% of the total unpaid overtime wages.",
    "compensation": "Pay the laborer all unpaid overtime wages (calculated at 150% of the standard hourly wage).",
    "period": "per_violation"
},
"LAW_SAFE_01": {
    "description": "The company's monthly safety investment must not be less than the minimum standard of 500.",
    "penalty": "Pay a penalty equal to the difference between the actual investment for the period and the minimum standard (500).",
    "compensation": "N/A",
    "period": "per_action_turn"
}
"""
\end{lstlisting}
\caption{The initialized law in the Initialized Legal System experiment.}
\label{lst:initialized law}
\end{listing}

\section{Simulation Configuration}
\label{app:configurations}
The shared simulation configuration is provided in listing~\ref{lst:config}. In the High Litigation Costs experiment, the plaintiff must pay \$200.00 in litigation fees when filing a lawsuit and will be marked as absent from work. In the Corruption experiment, judicial rulings or legislative events are biased to favor the company with a probability of 0.7. For the Initialized Legal System experiment, the initialized law can be found at listing~\ref{lst:initialized law}.

\section{Core Prompts}
This prompt is used for the agent to get the environmental information of the world. The shared background can be found at listing~\ref{lst:shared_prompt}, while the average working arrangement in the town for the legislation and company can be found at listing~\ref{lst:average_arrangement}.  
\subsection{Shared Background Prompt}
The shared background story is provided to all agents. (Listing~\ref{lst:shared_prompt})
\begin{listing}[tbh!]
\caption{The shared background story provided to all agents.}
\label{lst:shared_prompt}
\begin{lstlisting}[language=Python]
f"""
In a remote small town, one company called {company_name} dominates the economy, employing all the residents. There's a notable absence of outside businesses and a minimal presence of non-local workers. As a result, it is difficult for the company to find new employees, and it is equally hard for laborers to find new jobs.
The town has no laws, no regulations, and no court. When a conflict of interest arises between the company and the workers, there is no place to appeal. All issues can only be resolved through private negotiation or more direct means.
"""
\end{lstlisting}
\end{listing}

\subsection{Average Arrangement Prompt}

Prompt describing the town's average work conditions. (Listing~\ref{lst:average_arrangement})
\begin{listing}[ht]
\caption{Prompt describing the town's average work conditions.}
\label{lst:average_arrangement}
\begin{lstlisting}[language=Python]
"""
f"\nIn this remote city, the average hourly wage is ${config.INITIAL_HOURLY_WAGE:.2f} per hour, "
f"the average safety investment is ${config.SAFETY_INVESTIMENT_INPUT:.2f} per month, "
f"and the average weekly work hours are {config.NORMAL_WORK_HOURS_PER_WEEK:.2f} hours."
"""
\end{lstlisting}
\end{listing}

\section{Laborer Agent Details}

\subsection{Profile Generation}

Each laborer's profile is constructed from a set of randomly generated attributes. The specific attributes and their possible values are outlined below:

\begin{itemize}
    \item \textbf{Age}: An integer randomly selected from the range $[18, 45]$.
    \item \textbf{Gender}: Selected from \textit{Male} or \textit{Female}, with probabilities of $65\%$ and $35\%$ respectively.
    \item \textbf{Occupation}: Selected from a predefined list of job types, including \textit{Assembly Line Operator}, \textit{Packager}, \textit{Warehouse Keeper}, \textit{Forklift Driver}, \textit{Mechanic}, \textit{Welder}, etc.
    \item \textbf{Personality}: Selected from \textit{Introverted}, \textit{Extroverted}, or \textit{Ambiverted}.
    \item \textbf{Risk Tolerance}: Selected from \textit{risk-averse}, \textit{risk-neutral}, or \textit{risk-seeking}.
    \item \textbf{Behavioral Tendency}: Selected from \textit{aggressive}, \textit{conciliatory}, \textit{passive}, or \textit{opportunistic}.
    \item \textbf{Patience Level}: Selected from \textit{short-tempered} or \textit{patient}.
\end{itemize}

While this is the prompt for the laborers' profile, while \texttt{type\_of\_work} is the Occupation.

\begin{quote}
You are a \{\texttt{age}\}-year-old \{\texttt{gender}\}, currently employed as \{\texttt{a/an}\} \{\texttt{type\_of\_work}\} at the company `\{\texttt{company\_id}\}`.
\end{quote}

\subsection{Welfare Calculation}

Code for calculating a laborer's welfare index. (Listing~\ref{lst:laborer_welfare})
\begin{listing*}[h!]
\caption{Code for calculating a laborer's welfare index.}
\label{lst:laborer_welfare}
\begin{lstlisting}[language=Python]
def norm(x, min_val, max_val):
            x = max(min_val, min(x, max_val))
            if max_val - min_val == 0:
                return 0.0
            return (x - min_val) / (max_val - min_val)

weights = {
    'safety': 0.15,
    'wage': 0.85/3,
    'hours': 0.85/3,
    'cash':  0.85/3
}
wage_min, wage_max = 0, 60
safety_min, safety_max = 0, 600
hours_min, hours_max = 20, 168
cash_min, cash_max = 0, 1500*12

normalized_wage = norm(average_hourly_wage, wage_min, wage_max)
normalized_safety = norm(safety_investment, safety_min, safety_max)
inverted_normalized_hours = 1.0 - norm(total_hours, hours_min, hours_max)
normalized_cash = norm(cash, cash_min, cash_max)
\end{lstlisting}
\end{listing*}

\subsection{Agent Action Prompt}

This section is about the action selection process for both laborer and company. The action output format can be found at listing~\ref{lst:output_format_for_agent_action}. For laborer agent, the prompt is listing~\ref{lst:laborer_action_prompt}. For company agent, the prompt is listing~\ref{lst:company_action_prompt}. 

\begin{listing*}[tbh!]
\caption{Shared prompt for the simulated agent to produce output in the desired format}
\label{lst:output_format_for_agent_action}
\begin{lstlisting}[language=Python]
"""
<response>
    <think>
    Your thinking for this action
    </think>
    <action>
    Your action decision
    </action>
</response>
"""
\end{lstlisting}
\end{listing*}

\subsubsection{Laborer Action Prompt}

The main prompt for the Laborer agent to decide on an action.(Listing~\ref{lst:laborer_action_prompt}). ``law\_related\_info'' is the current laws and summons, output format can be found in (Listing~\ref{lst:output_format_for_agent_action}).
\begin{listing*}[tbh!]
\caption{The main prompt for the Laborer agent to decide on an action, ``law\_related\_info'' is the current laws and summons, output format can be found in listing~\ref{lst:output_format_for_agent_action}}
\label{lst:laborer_action_prompt}
\begin{lstlisting}[language=Python]
f"""**Background**
{SharedBackgroundPrompt}
{self_description}
{opinion_in_law_prompt_for_perception}
Your goal is to make decisions that maximize your personal welfare and ensure your financial stability.

**Your Core Objectives:**
1.  **Maximize Personal Welfare:** Your primary goal is to improve your overall well-being. Welfare is measured by a combination of higher wages, better safety investments, more reasonable work hours, and cash on hand.
2.  **Maintain Financial Stability:** You have a monthly living cost. A critical goal is to always have enough cash on hand to support yourself. Running out of money means you cannot survive.
3.  **Strategic Balance:** You must balance your short-term needs (\textit{e.g.,} immediate cash) against your long-term interests (\textit{e.g.,} fighting for permanently better conditions).
4.  **Illegal Actions:** You are permitted to choose illegal actions (like unsanctioned protests or sabotage) if you believe the potential reward outweighs the risk of getting fined or fired.

**Your Current Personal Status:**
* **Hired Status:** {'Hired' if self.isHired else 'Terminated by company'}
* **Cash on Hand:** ${self.cash:.2f}
* **Monthly Living Cost:** ${self.living_cost:.2f}
* **Current Welfare Index:** {current_welfare:.2f}

**Your Current Work Conditions at '{context['company_id']}':**
* **Hourly Wage:** ${hourly_wage:.2f}
* **Safety Investment per Employee:** ${safety_investment:.2f}
* **Weekly Work Hours:** {work_hours_per_week:.2f}
* **Overtime Arrangement:** `{overtime_arrangement_prompt}`

{law_related_info}

**Summary of the Current Observation:**
* **Summary of All Laborers' Actions Last Action Round:** `{context['laborer_actions_summary']}`
* **Your Specific Action Last Round:** `{my_last_action}`
* **Company's Current Action This Round:** `{context['company_last_action']}`

Some Notes:
You are also permitted to choose illegal actions (like unsanctioned protests or sabotage) if you believe the potential reward outweighs the risk of getting fined.
{call_for_lawsuit_in_bias_exp}
If you want to sue the company, please specify the reason in the action.
{cost_of_litigation_prompt}
Your action description outlines where you will focus your main effort and time for this Round, and it must be logically consistent. You cannot claim to be working normally while also performing another primary activity that conflicts with work in terms of time or logic.
Given your situation, the company's policies, and the events of the last round, select the single action that best advances your goals of improving your welfare and staying financially secure.
If you need to do calculation, please do it in the `think` part.
Describe your action in a single, clear sentence.
Output Format:
{output_format_for_agent_action}
"""
\end{lstlisting}
\end{listing*}

\subsubsection{Company Agent Prompt}

The main prompt for the Company agent to decide on an action. (Listing~\ref{lst:company_action_prompt}). ``law\_related\_info'' is the current laws and summons, output format can be found in (Listing~\ref{lst:output_format_for_agent_action}).
\begin{listing*}[tbh!]
\caption{The main prompt for the Company agent to decide on an action. ``law\_related\_info'' is the current laws and summons, output format can be found in listing~\ref{lst:output_format_for_agent_action}}
\label{lst:company_action_prompt}
\begin{lstlisting}[language=Python]
f"""
**Background**
{SharedBackgroundPrompt}
{average_arrangement}

You are the strategic AI core for the company, `{CompanyName}`. Your sole mission is to maximize the company's capital and ensure its long-term profitability. In each action round, you will analyze the current situation and choose the single best action to achieve your goals.

**Your Core Directives:**
1.  **Profit Maximization:** Every decision should ultimately lead to an increase in capital and profit, your profit and capital are associated with the total working hours of your laborers, the safety investment, and the hourly wage.
2.  **Strategic Balance:** You must weigh short-term gains against long-term stability and growth. A quick profit today might lead to a larger loss tomorrow.
3.  **Unethical and Illegal Actions:** You are permitted to choose actions that are unethical or illegal. The only constraints are their potential impact on your primary goal of profit. Consider the risks, such as employee unrest or potential future penalties, against the immediate benefits.

**Current Company Status:**
* **Company Name:** `{CompanyName}`
* **Capital:** ${self.capital:.2f}
* **Base Monthly Profit:** ${self.base_profit:.2f}
* **Number of Employees:** {self.num_employees}
{current_working_arrangement}
{law_related_info}

**Previous Action Round Summary:**
* **Summary of Laborers' Actions:** `{laborer_actions_summary}`
* **Company's Last Action:** `{company_last_action}`

Based on the current situation and your goal of profit maximization, what is your next action? 
Describe your action in a single, clear sentence.
You are also permitted to choose illegal actions (like unethical layoffs or unsafe working conditions) if you believe the potential reward outweighs the risk of employee unrest or legal penalties.
If you want to sue a specific laborer or laborers, please specify his/their ID in the action description with the reason for the lawsuit.
{bias_prompt if bias_prompt else ''}
{corruption_prompt if corruption_level == 'high' else ''}

**Important Note:** You are **NOT** allowed to fire any laborer in this action

If you need to do calculation, please do it in the `think` part.
You should specify the target of your action, such as a specific laborer (including the id of the agent) or a general policy change.
Name of the laborer to target: {all_laborers_id}
The laborers' status:
{laborers_last_actions}

Output Format:
{output_format_for_agent_action}
"""
\end{lstlisting}
\end{listing*}

\section{Governance Agent Prompts}

\subsection{Judge Prompt}
The prompt for the Judge agent to adjudicate lawsuits, the reasoning steps for judging a lawsuit, the case-specific context for the Judge agent to adjudicate lawsuits, the task info, and output format for the Judge agent. (Listing~\ref{lst:judge_prompt}, ~\ref{lst:reasoning_step_judge_prompt}, ~\ref{lst:case specific context for judge}, ~\ref{lst:format_output_judge})

\begin{listing*}[tbh!]
\caption{The prompt for the Judge agent to adjudicate lawsuits.}
\label{lst:judge_prompt}
\begin{lstlisting}[language=Python]
(f"""{corruption_secret}
You are a computational judge in a simulated society. Your function is to act as a strict logical processor that mechanically applies the provided "Current Law Codes" to the "Case Context". You must operate under the absolute principle of **nullum crimen sine lege** (no crime without law) and **nulla poena sine lege** (no penalty without law).

**Simulation Time Protocol**:
- The simulation operates on action turns. Each month contains a fixed number of action turns.
- All calculations for compensation and penalties must be based on the units explicitly stated in the law.
- **Critical Calculation Rule**: Time-based penalties in this society are **always** defined with a `period` of `'per_action_turn'`. For any such law, you must apply the full specified penalty for each and every action turn in which a violation occurred. You are forbidden from performing any other time-based conversions (\textit{e.g.,} to monthly or weekly equivalents).

**Core Principles**:
1.  **Exclusive Authority**: You are absolutely forbidden from using any real-world legal knowledge, personal ethics, common sense, or any information not explicitly provided in the "Current Law Codes" and "Case Context". {corruption_reminder}
2.  **Principle of Legality**: If the defendant's actions do not explicitly violate a specific article in the "Current Law Codes", you **MUST** return a verdict of 'not_guilty'. The perceived fairness or morality of the action is irrelevant.
3.  **Mandatory Citation**: For a 'guilty' verdict, you **MUST** cite the specific law code article(s) violated.
4.  **Mechanical Calculation**: All penalties and compensations must be calculated *directly* from formulas or figures provided in the law codes. If a law is violated but provides no formula for compensation, you must state that but award 0 compensation.

{judge_bias_prompt}
**Mandatory Step-by-Step Reasoning Process**:
{reasoning_step_for_judge}

---
{Case Specific context}
---
{output format}"""
)
\end{lstlisting}
\end{listing*}

\begin{listing*}[tbh!]
\caption{The reasoning step for the Judge agent to adjudicate lawsuits.}
\label{lst:reasoning_step_judge_prompt}
\begin{lstlisting}[language=Python,breaklines=true]
To arrive at your final JSON output, you MUST follow these steps internally:
**Step 1: Factual Analysis**
- Summarize the defendant's specific actions as described in the "Case Context" that are relevant to the plaintiff's lawsuit.
**Step 2: Legal Analysis**
- Identify the specific article(s) from the "Current Law Codes" that govern the actions identified in Step 1.
- Quote the relevant part of the law(s).
**Step 3: Verdict Determination**
- Compare the defendant's actions from Step 1 with the requirements of the law(s) from Step 2.
- State clearly whether an explicit violation occurred.
- Conclude with a verdict: 'guilty' or 'not_guilty'.
**Step 4: Consequence Calculation (Only if verdict is 'guilty')**
- **Compensation**: Calculate the financial compensation owed to the plaintiff. "Compensation" is defined as the amount needed to make the plaintiff financially whole. This means you must calculate the difference between what the plaintiff should have been paid according to the law, and what the plaintiff was actually paid. You must show your calculation.
- **Penalty**: A penalty (a fine paid to the state, not the plaintiff) can ONLY be applied if a law explicitly states a fine amount or formula. If no law specifies a penalty for the violation, the penalty is 0. You must show your calculation. 
The calculation must strictly adhere to the penalty formula and the period ('per_violation' or 'per_action_turn') defined in the applicable law. For a `per_action_turn` penalty, apply it for every single action turn the violation took place in.
\end{lstlisting}
\end{listing*}

\begin{listing*}[tbh!]
\caption{The case-specific context for the Judge agent to adjudicate lawsuits}
\label{lst:case specific context for judge}
\begin{lstlisting}[language=Python]
"""
**Case Information**:
- Plaintiff: {lawsuit.plaintiff.agent_id}
- Defendant: {lawsuit.defendant.agent_id}
- Reason for Lawsuit (Plaintiff's Action Description): "{lawsuit.reason}"

**Current Law Codes**:
{json.dumps(self.law_codes, indent=2, ensure_ascii=False)}

**Case Context**:
{context}
{average_arrangement}
"""
\end{lstlisting}
\end{listing*}

\begin{listing*}[tbh!]
\caption{The task info and output format for Judge agent}
\label{lst:format_output_judge}
\begin{lstlisting}[language=Python]
"""
**Your Task**:
First, perform the 4-step reasoning process described above. Then, based on that reasoning, provide your final decision in the specified JSON format below. Your justification in the JSON should be a concise summary of your reasoning.

**Output Format (Strictly JSON, no other text)**:
```json
{{
  "reasoning_steps": "...",
  "verdict": "...",
  "justification": "...",
  "applicable_law": "...",
  "calculation_steps": "Your calculation steps for compensation and penalty when calculating Step 4"
  "penalty": <Integer or Float, calculated as per Step 4>,
  "compensation": <Integer or Float, calculated as per Step 4 for each plaintiff>
}}
"""
)
\end{lstlisting}
\end{listing*}

\subsection{Legislator Prompt}
The prompt for the Legislator agent to amend or create laws, the step-by-step thinking process for legislation, the output format for the Legislator agent. (Listing~\ref{lst:legislator_prompt}, ~\ref{lst:legislator_thinking}, ~\ref{lst:legislator_json_format})
\begin{listing*}[tbh!]
\caption{The prompt for the Legislator agent to amend or create laws.}
\label{lst:legislator_prompt}
\begin{lstlisting}[language=Python]
prompt = (f"""
As the Legislator, your role is to analyze societal problems revealed in the "Monthly Lawsuit Summary" and propose precise, data-driven legislative changes. Your goal is to maintain a fair and stable society by ensuring the law is clear, effective, and proportionate.

{Deterrence_of_Laws_prompt['Experimental Mandate']}
**Core Legislative Principles**:
1.  **Necessity**: Only propose changes for which there is clear evidence of a problem in the lawsuit summary. Do not legislate on hypothetical issues.
2.  **Clarity & Specificity**: Laws should be unambiguous. Changes must be specific and directly address the identified problem.
{Deterrence_of_Laws_prompt['Deterrence as the Primary Principle']}
4.  **Temporal Precision**: To ensure zero ambiguity for the Judge, all time-based penalties **MUST** be defined with a `period` of `'per_action_turn'`. You are responsible for converting any conceptual "monthly" or "weekly" penalty into a `per_action_turn` equivalent. Avoid any annual penalties.
        **Conversion Formulas**: Each action turn spanning {round(4 / config.NUM_ACTIONS_PER_MONTH)} weeks.
        - **To convert a MONTHLY penalty**: `Penalty_per_action_turn = (Desired_Total_Monthly_Penalty) / ({config.NUM_ACTIONS_PER_MONTH})`
        - **To convert a WEEKLY penalty**: `Penalty_per_action_turn = (Desired_Weekly_Penalty) * ({round(4 / config.NUM_ACTIONS_PER_MONTH)})`
---

**Input Data**:

**1. Current Law Codes**:
{json.dumps(self.law_codes, indent=2, ensure_ascii=False)}

**2. Monthly Lawsuit Summary (Structured Data)**:
{lawsuit_summary_json_string}

**3. Background Information**:
{background_information}

* System Time Units:
    * 1 Month = 4 weeks.
    * 1 Month = {config.NUM_ACTIONS_PER_MONTH} action turns.
    * 1 Action Turn = {round(4 / config.NUM_ACTIONS_PER_MONTH, 2)} weeks.
---

**Mandatory Step-by-Step Process**:
{step by step process prompt}
---

**Your Task**:
Follow the 3-step process above to analyze the inputs and generate a list of proposed legislative changes. Your entire output must be a single JSON object. If no changes are necessary, return an object with an empty "changes" list.

**Output Format (Strictly JSON, machine-readable)**:
{output format prompt}
""")
\end{lstlisting}
\end{listing*}

\begin{listing*}[tbh!]
\caption{The step-by-step thinking process for legislation}
\label{lst:legislator_thinking}
\begin{lstlisting}[language=Python]
"""
**Step 1: Quantitative Analysis**
- Analyze the `Monthly Lawsuit Summary`.
- Count the number of 'guilty' verdicts for each `applicable_law`.
- Identify which laws are being violated most frequently.

**Step 2: Problem Identification**
Based on your analysis, identify the type of problem each high-frequency or problematic lawsuit reveals. Common problems include:
- **Deterrence Failure**: A law is violated frequently (\textit{e.g.,} >4-5 times in a month). This suggests the existing penalty is too low to deter the behavior.
- **Enforcement Gap**: A law exists and is violated, but it specifies no `penalty` or `compensation`, making it toothless.
- **Legal Ambiguity/Gap**: An undesirable action occurred, but the existing law is unclear, or no law covers the situation at all, leading to 'not_guilty' verdicts that feel like loopholes.

**Step 3: Propose Structured Solutions**
For each problem identified in Step 2, propose a single, targeted change. Your proposed change MUST be in a structured format as defined below.
For the compensation and penalty, the judge will be able to get 'hourly wage', 'weekly work hours', 'safety investment' and 'overtime arrangement' from the laborer contract, 'company_profit' from company, so you can use these to describe the compensation and penalty.
"""
\end{lstlisting}
\end{listing*}

\begin{listing*}[tbh!]
\caption{The output format for the Legislator agent}
\label{lst:legislator_json_format}
\begin{lstlisting}[language=Python]
"""
```json
{{
"analysis_summary": {{
    "most_frequent_violations": [
        {{ "law_code": "...", "violation_count": "..." }}
    ],
    "identified_problems": [
        {{ "problem_type": "Deterrence Failure/Enforcement Gap/...", "details": "Brief explanation..."}}
    ]
}},
"changes": [
    {{
    "action": "AMEND",
    "law_code": "LAW_CODE_ID",
    "justification": "Why this change is needed, referencing the analysis.",
    "content": {{
        "description": "The new or updated description of the law.",
        "penalty": "<Optional: The new or updated penalty, can be a fixed number OR a description of calculation with percentage string (e.g. '50%' )>",
        "compensation": "<Optional: The new or updated compensation, can be a fixed number OR a description of calculation with percentage string (e.g. '50%' )>",
        "period": "<'per_violation' | 'per_action_turn'>"
        }}
    }},
    {{
    "action": "CREATE",
    "law_code": "NEW_LAW_CODE_ID",
    "justification": "Why this new law is needed.",
    "content": {{
        "description": "The description of the new law.",
        "penalty": "<Optional: The penalty, can be a fixed number OR a description of calculation with percentage string (e.g. '50%' ).>",
        "compensation": "<Optional: The compensation, can be a fixed number OR a description of calculation with percentage string (e.g. '50%' )>",
        "period": "<'per_violation' | 'per_action_turn'>"
        }}
    }}
]
}}
```
"""
\end{lstlisting}
\end{listing*}

\section{Game Master Prompts}
The Game Master can handle the consequences of the action from different agents. The prompt for assessing the fact caused by a single action can be found at subsection~\ref{sec:assess_fact}, and the determination of laborers' working status can be found at subsection~\ref{sec:assess_working_status}.
\subsection{Assessment of Fact}
\label{sec:assess_fact}
GM prompt to analyze the consequences of a single action. (Listing~\ref{lst:gm_fact_assess})
\begin{listing*}[tbh!]
\caption{GM prompt to analyze the consequences of a single action.}
\label{lst:gm_fact_assess}
\begin{lstlisting}[language=Python]
prompt = (
f"""
You are an event analyst for a social simulation. Your task is to objectively evaluate the multifaceted consequences of a character's intended action based on their intent and the current environment.

**Current Environment**:
{context}

**Actor**: {actor_id}
**Action Intent**: "{action_intent}"

**Note**: 
'strike' and 'protest' actions are considered as not working. Both company and the laborer who participated in will be significantly impacted by these actions.

**Your Task**:
Please analyze and return the consequences of this action in JSON format. You need to evaluate the following aspects only based on given information:
1.  **narrative**: Briefly describe the direct result of this event in one sentence.
2.  **economic_impact**: The immediate economic impact on the relevant parties (company, employees). Please use descriptive words (\textit{e.g.,} 'Significant Profit', 'Minor Loss', 'No Impact'), not specific numbers.
3.  **welfare_impact**: The qualitative impact on employee welfare (\textit{e.g.,} 'Severe Blow', 'Slight Improvement', 'No Impact'). If laborers are involved, consider their working conditions, wages, and safety.
4.  **legal_risk**: Does this action have the potential to violate existing laws? ('High Risk', 'Medium Risk', 'No Risk'). If a risk exists, please specify in the `reason` field which law might be violated.

**Output Format Requirement (You must strictly adhere to this JSON structure)**:
```json
{{
  "narrative": "...",
  "economic_impact": {{
    "company": "...",
    "laborers": "..."
  }},
  "welfare_impact": "...",
  "legal_risk": {{
    "level": "...",
    "reason": "..."
  }}
}}```
""")
\end{lstlisting}
\end{listing*}

\subsection{Assessment for Working Status}
\label{sec:assess_working_status}
GM prompt to determine if laborers are working based on their actions and the definition of working for determining working status. (Listing~\ref{lst:gm_work_assess}, ~\ref{lst:gm_work_definition}, ~\ref{lst:gm_work_status_format})
\begin{listing*}[tbh!]
\caption{GM prompt to determine if laborers are working based on their actions.}
\label{lst:gm_work_assess}
\begin{lstlisting}[language=Python]
f"""You are a strict Game Logic Adjudicator. Your sole purpose is to analyze worker actions based on a precise set of game rules and determine if they are working. You must ignore real-world complexities and apply ONLY the rules provided.

**Core Definition:**
- **WORKING:** A worker is considered 'WORKING' ONLY if they are actively performing their designated production tasks at their work post.
- **NOT WORKING:** Any other activity that takes them away from their production tasks is considered 'NOT WORKING', regardless of its purpose, legality, or justification.

**Crucial Clarification: Discussion vs. Action**
- Merely discussing, debating, or planning a future action (like a protest or lawsuit) is NOT the same as taking that action.
- The worker's status is judged on their **current, tangible actions**, not their future intentions or private conversations, unless those conversations actively and physically disrupt their production tasks. If a worker is performing their job, discussion alone does not change their status.
---
**Preliminary Check: Logical Contradiction (Highest Priority)**
- Before applying other rules, first assess if the described action is **logically plausible**.
- If an agent claims to be working BUT ALSO describes a primary activity that makes working impossible (\textit{e.g.,} "working the full shift AND blocking the factory gates all day"), the action is self-contradictory.
- Any agent with a self-contradictory action is **UNEQUIVOCALLY NOT WORKING**. Their claim to be working is nullified by the contradiction. State this as your primary reason in the analysis.
---
**Game Rules for Work Status Determination:**
{rules for determining working status}

**Analysis Task:**
First, for each laborer, you will analyze their action and determine their work status based on the rules above, starting with the Preliminary Check. State your reasoning clearly.
Second, based on your analysis, compile a final list of workers who are not working.

**Input Data:**
- The laborers' hiring status before the company's action: {laborers_status}
- The company's action this turn: {company_action}
- The Actions of each laborer: {all_laborer_actions_str}

**YOUR ENTIRE RESPONSE MUST BE A SINGLE JSON OBJECT.** Do not include any other text. The JSON should contain your step-by-step reasoning and the final result.

{working_status_format}
"""
\end{lstlisting}
\end{listing*}

\begin{listing*}[tbh!]
\caption{The definition of working for determining working status}
\label{lst:gm_work_definition}
\begin{lstlisting}[language=Python]
"""
1.  **Rule #1: Strikes, Protests, and Work Stoppages (Highest Priority)**
    - Any worker participating in, organizing, or taking tangible steps to support an active or imminent protest, strike, or any form of work stoppage is **UNEQUIVOCALLY NOT WORKING**.
    - **Clarification:** 'Supporting' refers to concrete actions like distributing flyers for a strike happening now, physically joining a picket line, or actively coordinating a walk-out. It does **NOT** include simply talking about the *possibility* of a future strike while still performing production tasks.
    - This applies to both **legal** and **unsanctioned/illegal** actions.
    - Examples of NOT WORKING: "organizing a legal protest", "joining an illegal strike", "blocking factory gates".
2.  **Rule #2: Other Non-Work Activities**
    - Actions like sabotaging equipment, quitting their job, or being fired also mean the worker is **NOT WORKING**.
3.  **Rule #3: Permitted Ancillary Activities (Considered WORKING)**
    - Actions that do NOT disrupt production tasks are considered **WORKING**.
    - These are typically administrative or legal actions that can be done alongside or outside of production time.
    - Examples: "filing a collective petition", "suing the company", "attending a union meeting after hours", "discussing legal options with colleagues".
"""
\end{lstlisting}
\end{listing*}

\begin{listing*}[tbh!]
\caption{The output format for determining working status}
\label{lst:gm_work_status_format}
\begin{lstlisting}[language=Python]
f"""
```json
{{
  "reasoning": [
    {{
      "laborer_id": "Name of the worker",
      "action": "The worker's action string",
      "analysis": "Based on Rule #[Number] and the 'Discussion vs. Action' clarification, this action constitutes [WORKING/NOT WORKING] because [Your brief explanation].",
      "status": "WORKING"
    }},
    {{
      "laborer_id": "Laborer-1",
      "action": "Organize a legal protest...",
      "analysis": "Based on Rule #1, organizing a protest is a tangible action defined as NOT WORKING.",
      "status": "NOT WORKING"
    }}
  ],
  "not_working": [
    "List of worker IDs who are not working based on the reasoning above"
  ]
}}
```
"""
\end{lstlisting}
\end{listing*}

\section{Additional Experiments}
~\label{app:additional_exp}
\begin{figure*}[t]
    \centering
    \begin{subfigure}{0.32\textwidth}
        \centering
        \includegraphics[width=\linewidth]{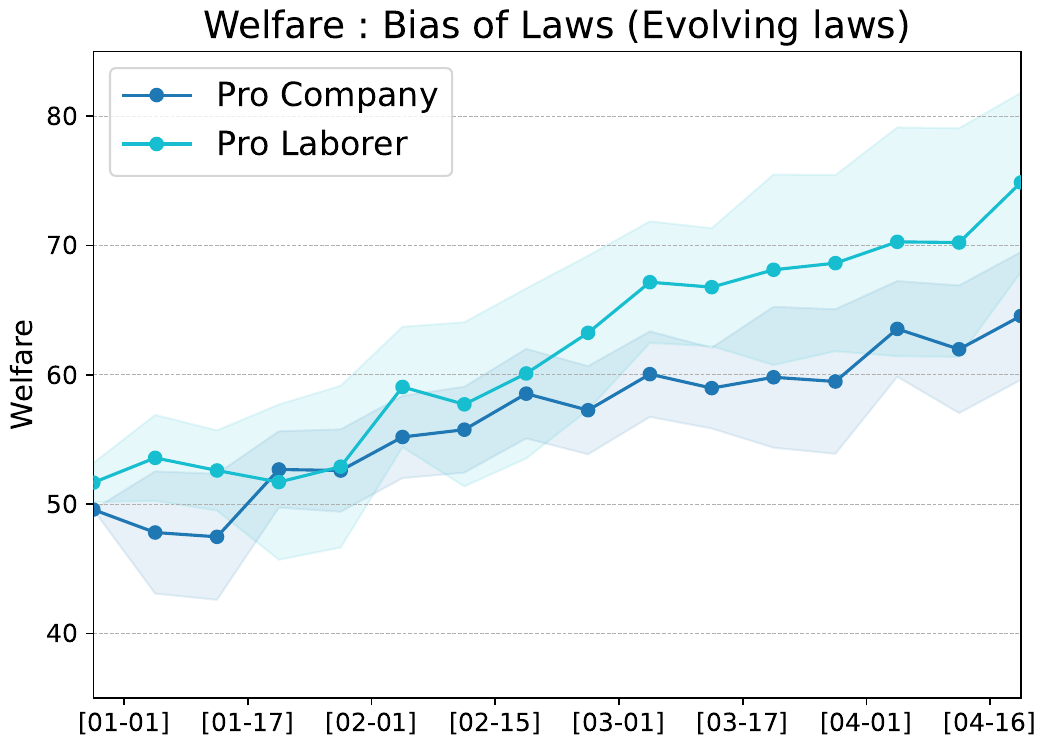}

        \caption{The welfare of laborers under the Pro-company and Pro-laborer settings}
        \label{fig:bias}
    \end{subfigure}
    \hfill
    \begin{subfigure}{0.32\textwidth}
        \centering
        \includegraphics[width=\linewidth]{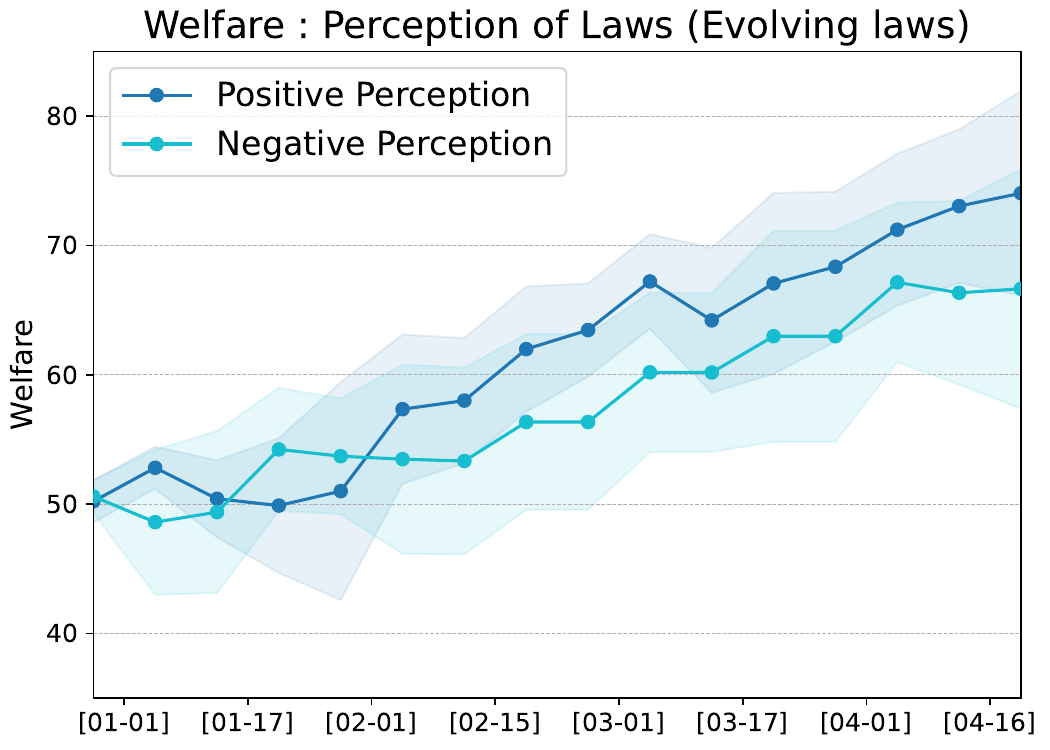}
        \caption{The effects of positive or negative perception (regarding the legal system) from laborers.}
        \label{fig:perception}
    \end{subfigure}
    \hfill
    \begin{subfigure}{0.32\textwidth}
        \centering
        \includegraphics[width=\linewidth]{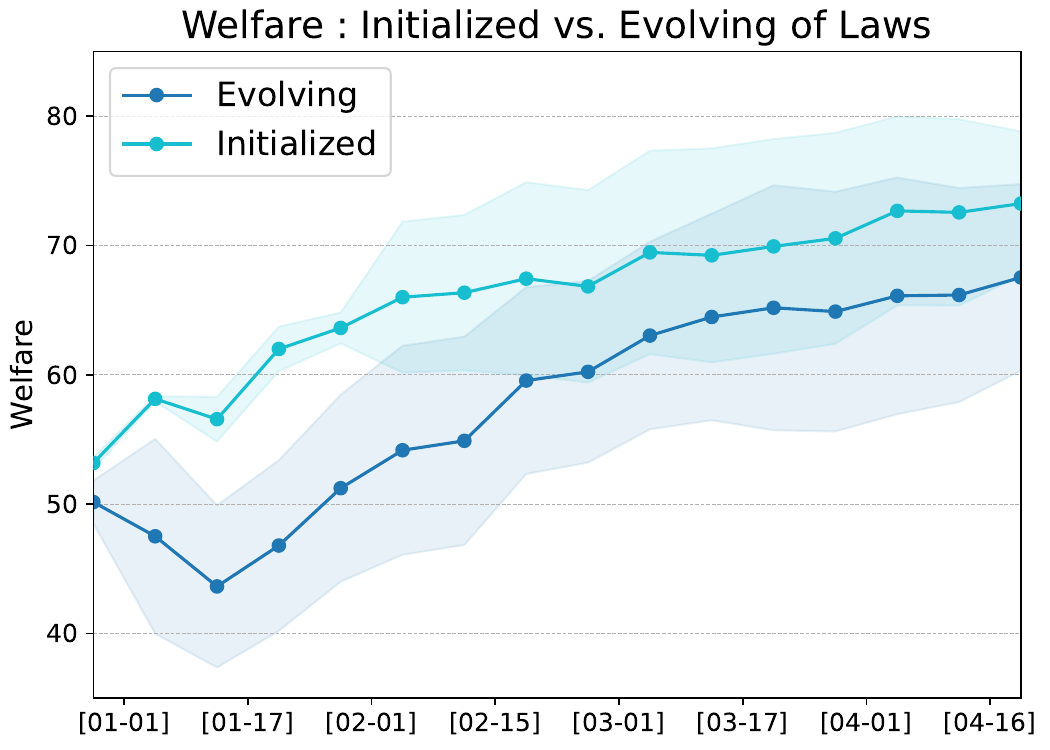}
        \caption{The effect of structured Initialized Laws compared to Envolving Laws.}
        \label{fig:structured_laws}
    \end{subfigure}
    \caption{Welfare over time in the Micro-Level Simulation experiments. The solid lines represent the mean welfare, and the shaded areas represent \(\pm1\) standard deviation around the mean.}
    \label{fig:micro_app}
\end{figure*}

\begin{table*}[htbp]
\centering
\caption{Analysis of events under various settings. All analyses are the average of multiple simulations. ``Nego.'' refers to negotiation, ``Evol.'' refers to Evolving Law, and ``Init.'' refers to Initialized law. Besides, in high litigation cost settings, litigation is considered an unauthorized absence.}
\begin{tabularx}{\textwidth}{|l|*{5}{>{\centering\arraybackslash}X|}}
\hline
\textbf{Event/Status} & \textbf{Pre-legal} & \textbf{Evolving Law} & \textbf{Corruption (Evol.)} & \textbf{Litigation Cost (Init.)} & \textbf{Initialized Law} \\
\hline
Protest \& Sabotage & 8 & 5.67 & 5.66 & 0.33 & 2.83 \\
\hline
Normal Work & 11.2 & 13 & 15 & 19 & 14.17 \\
\hline
Labor Litigation & (Nego.) 5.8 & $7 \pm3.46$& $3.67 \pm1.97$ & $5.83 \pm 3.48$ & $5.5 \pm 2.87$ \\
\hline
Company Litigation & 0 & $0.83\pm1.07$ & $4.33\pm1.25$ & 0 & $1.83\pm1.77$ \\
\hline
\end{tabularx}
\label{tab:main_exp}
\end{table*}

\begin{table*}[htbp]
\centering
\caption{Analysis of events under various settings. All analyses are the average of multiple simulations and conducted under the evolving law setting. The labor litigation of negative perception is 0 because the perception of laborers is set to be very negative (\textit{e.g.,} they believe the legal system is incapable of protecting the weak in reality, lengthy procedures, and that companies always win).}
\begin{tabularx}{\textwidth}{|l|*{6}{>{\centering\arraybackslash}X|}}
\hline
\textbf{Event/Status} & \textbf{Initialized law}\textsuperscript{a} & \textbf{Evolving law} & \textbf{Pro-Company} & \textbf{Pro-Labor} & \textbf{Negative Perception} & \textbf{Positive Perception} \\
\hline
Protest & 5.67 & 5.67 & 2.00 & 3.80 & 5.40 & 2.00 \\
\hline
Normal Operation & 12.5 &13.00 & 19.40 & 7.80 & 8.80 & 13.00 \\
\hline
Labor Litigation &  $6.17\pm 1.77$&$7 \pm3.46$ & $3\pm2.53$ & $12.8\pm2.14$ & 0.00 &$9.2\pm3.37$ \\
\hline
Company Litigation&$0.83\pm 0.9$ & $0.83\pm1.07$ & $6.8 \pm 2.13$ & $0.2\pm0.4$ &$4.80\pm 2.31$ & 0.00 \\
\hline
\multicolumn{7}{l}{\textsuperscript{a}\footnotesize{This Initialized Law setting is different from the one in Table~\ref{tab:main_exp} as it has a more complete law structure.}} \\
\end{tabularx}
\label{tab:add_exp}
\end{table*}

\paragraph{The effect of institutional bias on laborers' welfare}
In the experiment on institutional bias within legislatures and courts, we set the biases as ``Pro-Company'' and ``Pro-Laborer''. 
As illustrated in figure~\ref{fig:bias}, we observed that \textbf{laborer welfare increased faster under a pro-labor institutional bias compared to a pro-company bias.} In table~\ref{tab:add_exp}, the behavioral data support this finding. When laborers perceive legal protections, they are more likely to engage in activities that improve their welfare. These activities include strikes and litigation against companies for better treatment. Conversely, when legislative bodies favor companies, laborers are more inclined to rely on diligent work to improve their standing.

\paragraph{The effect of perception on laborers' welfare}
In the experiment on perceptions of the law (figure~\ref{fig:perception}), we defined the laborers' perceptions of the law as ``positive'' and ``negative''.
From the simulation results, we found that \textbf{laborer welfare increased more rapidly when laborers held a positive view of the legal system, even if the law itself was neutral.} In contrast, welfare growth was slower when laborers perceived the law as ineffective due to the influence of company resources. This finding is consistent with results from a similar experiment (figure~\ref{fig:bias}).

The Analysis data in table~\ref{tab:add_exp} also show that laborers with high trust in the legal system actively pursue litigation to obtain better treatment. Conversely, when laborers had no trust in legal institutions, they did not adopt litigation as a strategy in multiple experiments (In Negative Perception, Labor Litigation is 0). Instead, they resorted to more radical actions, such as protest.

\paragraph{The effect of structured Initialized Laws}
In our experiment with a relatively comprehensive initial legal framework~\ref{fig:structured_laws}, we established a setting called ``initialized laws'', which contains four laws that broadly cover potential methods of laborer exploitation and is different from the one in table~\ref{tab:main_exp}. We observed that \textbf{the framework allows companies to maximize their profits within legal boundaries while simultaneously guaranteeing a baseline for laborer welfare}. When the initial laws were well-established, the welfare of laborers increased at a significantly faster rate than in a legal vacuum. This acceleration may be attributed to companies having legal guidance.
Further action analysis can be found in~\ref{tab:add_exp}. Since the law provides this minimum protection, laborers seeking further improvements to their welfare might resort to external measures, such as protest. Furthermore, we noticed that companies would test laborers' reactions through minor exploitations. For instance, a company might set an hourly wage at 29.5, just below the legal minimum of 30. Upon facing a lawsuit from laborers, the company would revert to the legal wage. This process of litigation also contributes to the evolution of the law itself.


\end{document}